\documentclass{ieeeaccess}
\usepackage{placeins}
\usepackage{booktabs}
\usepackage{siunitx}
\usepackage{mathtools}
\DeclarePairedDelimiter\ceil{\lceil}{\rceil}
\usepackage{enumitem}
\usepackage{multirow}
\usepackage{floatrow}
\usepackage{float}
\usepackage{hyperref}
\usepackage[utf8]{inputenc}
\usepackage{fourier} 
\usepackage{makecell}
\usepackage{tabularx}
\usepackage{ragged2e}
\usepackage{cite}
\usepackage{amsmath,amssymb,amsfonts}
\usepackage{algorithmic}
\usepackage{graphicx}
\usepackage{textcomp}
\def\BibTeX{{\rm B\kern-.05em{\sc i\kern-.025em b}\kern-.08em
    T\kern-.1667em\lower.7ex\hbox{E}\kern-.125emX}}
\begin{document}
\history{Date of publication xxxx 00, 0000, date of current version xxxx 00, 0000.}
\doi{10.1109/ACCESS.2017.DOI}

\title{Deep Graph Generators: A Survey}
\author{\uppercase{Faezeh Faez}\authorrefmark{1},
\uppercase{Yassaman Ommi}\authorrefmark{2}, \uppercase{Mahdieh Soleymani Baghshah}\authorrefmark{1}, \uppercase{and Hamid R. Rabiee}\authorrefmark{1},
\IEEEmembership{Senior Member, IEEE}}
\address[1]{Department of Computer Engineering, Sharif University of Technology, Tehran, Iran}
\address[2]{Department of Mathematics and Computer Science, Amirkabir University of Technology, Tehran, Iran}

\markboth
{F. Faez \headeretal: Deep Graph Generators: A Survey}
{F. Faez \headeretal: Deep Graph Generators: A Survey}

\corresp{Corresponding authors: Hamid R. Rabbiee and Mahdieh Soleymani Baghshah (e-mails: rabiee@sharif.edu , soleymani@sharif.edu).}

\begin{abstract}
Deep generative models have achieved great success in areas such as image, speech, and natural language processing in the past few years. Thanks to the advances in graph-based deep learning, and in particular graph representation learning, deep graph generation methods have recently emerged with new applications ranging from discovering novel molecular structures to modeling social networks. This paper conducts a comprehensive survey on deep learning-based graph generation approaches and classifies them into five broad categories, namely, autoregressive, autoencoder-based, RL-based, adversarial, and flow-based graph generators, providing the readers a detailed description of the methods in each class. We also present publicly available source codes, commonly used datasets, and the most widely utilized evaluation metrics. Finally, we highlight the existing challenges and discuss future research directions.
\end{abstract}


\titlepgskip=-15pt

\maketitle

\section{Introduction}
\begin{table*}[!tp]
\begin{center}
\small
 \begin{tabular}{||m{0.2\textwidth} m{0.49\textwidth} m{0.23\textwidth}||}
 \hline
 Category & Key Characteristic & Publications\\ [0.5ex] 
 \hline\hline
 Autoregressive DGGs & Adopting a sequential generation strategy, either node-by-node or edge-by-edge & \cite{li2018multi, you2018graphrnn, popova2019molecularrnn, bacciu2019graph, bacciu2020edge, goyal2020graphgen, liao2019efficient, kawai2019scalable, fan2020attention, li2018learning, dai2020scalable, shi2020graphaf, shah2020auto, jin2020hierarchical, bradshaw2019model, liu2018constrained, assouel2018defactor, su2019graph, lim2020scaffold, you2018graph, shi2019reinforced, karimi2020network, khemchandani2020deepgraphmolgen, trivedi2020graphopt, xu2020reinforced,  ahn2020guiding}\\ 
 \hline
 Autoencoder-Based DGGs & Making the generation process dependent on latent space variables & \cite{kipf2016variational, simonovsky2018graphvae, flam2020graph, ma2018constrained, grover2019graphite, guo2020interpretable, li2020dirichlet, jin2018junction, jin2020hierarchical, kajino2019molecular, bradshaw2019model,liu2018constrained, assouel2018defactor, samanta2019nevae, liu2019graph, yang2019conditional, jin2018learning, su2019graph, lim2020scaffold}\\
 \hline
 RL-Based DGGs & Utilizing reinforcement learning algorithms to induce desired properties in the generated graphs & \cite{popova2019molecularrnn, you2018graph, shi2019reinforced, karimi2020network, khemchandani2020deepgraphmolgen, trivedi2020graphopt, ahn2020guiding, de2018molgan, xu2020reinforced}\\
 \hline
 Adversarial DGGs & Employing generative adversarial networks (GANs) \cite{goodfellow2014generative} to generate graph structures & \cite{bojchevski2018netgan, gamage2020multi, tann2020shadowcast, you2018graph, karimi2020network, de2018molgan, yang2019conditional, yang2020learn, jin2018learning, maziarka2020mol, zhou2019misc}\\ 
 \hline
 Flow-based DGGs & Learning a mapping from the complicated graph distribution into a distribution mostly modeled as a Gaussian for calculating the exact data likelihood & \cite{shi2020graphaf, shah2020auto, liu2019graph, madhawa2019graphnvp}\\
 [1ex] 
 \hline
\end{tabular}
\caption{\label{tab:categorization} Categorization,  Key Characteristic, and Representative Publications among Deep Graph Generators}
\end{center}
\end{table*}
Recently, with the rapid development of data collection and storage technologies, an increasing amount of data that needs to be processed is available. In many research areas, including biology, chemistry, pharmacy, social networks, and knowledge graphs, there exist some relationships between data entities that, if taken into account, more valuable features can be extracted, yielding more accurate predictions. Using graph data structure is a common way to represent such data, and therefore graph analysis research has attracted considerable attention.\par
In the past few years, graph-related studies have made significant progress, which mainly focus on graph representation learning \cite{kipf2016semi, hamilton2017inductive, xu2018powerful, cotta2020unsupervised, ramezani2020gcn} but also include other problems like graph matching \cite{li2019graph, fey2019deep}, adversarial attack and defense on graph-based neural networks \cite{zugner2018adversarial, zugner2019adversarial}, and graph attention networks \cite{velivckovic2017graph, kosaraju2019social}. Graph generation is also another research line aiming to generate new graph structures with some desired properties, which dates back to 1960 \cite{erdHos1960evolution} and is followed by several other approaches \cite{watts1998collective, albert2002statistical, holland1983stochastic, leskovec2010kronecker}.  However, the early methods generally use hand-engineered processes to create graphs with predefined statistical properties and, despite their simplicity, are not capable enough to capture complicated graph dependencies.\par
Thanks to the recent successes of deep learning techniques and algorithms, deep generative models, which aim to generate novel samples from a similar distribution as the training data, have received a lot of attention in various data domains such as image\cite{radford2015unsupervised, yan2016attribute2image}, text\cite{zhang2017adversarial, vae22}, and speech\cite{kaneko2017sequence, gao2018voice}. Subsequently, studies related to deep learning-based graph generators have started a little later, which, unlike the traditional approaches, can directly learn from data and eliminate the need for using hand-designed procedures. Therefore, there are apparent horizons in this research area, with applications ranging from discovering new molecular structures to modeling social networks.\par
So far, several surveys have reviewed deep graph-related approaches such as those mainly focusing on graph representation learning methods \cite{zhang2020deep, wu2020comprehensive, zhou2018graph, cao2020comprehensive, bacciu2020gentle}, graph attention models \cite{lee2019attention}, attack and defense techniques on graph data \cite{sun2018adversarial}, and graph matching approaches \cite{ma2019deep, yan2020learning}. Although most of these surveys have made a passing reference to the modern graph generation approaches, which we refer to as Deep Graph Generators (DGGs), this field requires individual attention due to its value and expanding development.\par 
In this paper, we conduct a survey on DGGs in order to exclusively review and categorize these methods and their applications. To this end, we first divide the existing approaches into five broad categories, namely, autoregressive DGGs, autoencoder-based DGGs, RL-based DGGs, adversarial DGGs, and flow-based DGGs, providing the readers with detailed descriptions of the methods in each category and comparing them from different aspects. This categorization is either based on the model architectures, adopted generation strategies, or optimization objectives and the categories may sometimes overlap so that a method can belong to more than one category. Table \ref{tab:categorization} summarizes the main characteristics of these categories, along with the most prominent approaches belonging to each of them.\par 
The rest of this article is organized as follows. Section \ref{sec:notations} briefly summarizes notations used in this survey and formulates the problem of deep graph generation. Sections \ref{sec:autoregressive} to \ref{sec:flow} provide a detailed review of the existing DGGs in each of the five categories discussed above. Section \ref{sec:applications} classifies the current applications and suggest some potential future ones. Section \ref{sec:implementations} goes through implementation details by summarizing commonly used datasets, widely utilized evaluation metrics, and available source codes. Section \ref{sec:future} discusses future research directions. Finally, section \ref{sec:conclusion} concludes the survey.
\section{Notations and Problem Formulation}\label{sec:notations}
\begin{table}
\small
\caption{\label{tab:notations} COMMONLY USED NOTATIONS}
\centering
\begin{tabular}{p{2 cm} p{5.50cm}} \toprule
    {Notations} & {Descriptions} \\ \midrule
    G   & {A graph.}     \\ \midrule
    V   & {The node set of a graph.}     \\ \midrule
    {E}   & {The edge set of a graph.}     \\ \midrule
    {p(G)}  & {The graph data distribution.} \\ \midrule
    {n}  & {The number of nodes, n=|V|.} \\ \midrule
    {N}  & {The largest graph size in the dataset.} \\ \midrule
    {m}  & {The number of edges, m=|E|.} \\ \midrule
    {$\pi$}  & {A node ordering for a graph G.} \\ \midrule
    {$A^\pi$}  & {The adjacency matrix corresponding to a graph G under a node ordering $\pi$.} \\ \midrule
    {$S^\pi$}  & {The sequence corresponding to nodes of a graph under a node ordering $\pi$.} \\ \midrule
    {$S^{edge,\ \pi}$}  & {The sequence corresponding to edges of a graph.} \\ \midrule
    {$emb(.)$}  & {An embedding function.} \\ \midrule
    {$[x, y]$}  & {The concatenation of x and y.} \\ \midrule
    {$\mathcal{N}_u$}  & {The neighbor set of node $u$.} \\ \bottomrule
\end{tabular}
\end{table}
This section reviews the notations used in the survey and provides a problem formulation for generating a set of plausible graphs.
\subsection{Notations} 
 We represent a graph as $G=(V, E)$, where $V$ is the graph's node set, and $E$ denotes its edge set, with $|V|=n$ and $|E|=m$. $N$ also indicates the largest graph size in the dataset. There are $n!$ possible node orderings for the graph; thus, if we choose an ordering $\pi$, the graph can be represented by the corresponding adjacency matrix $A^\pi \in \mathbb{R}^{n\times n}$. Moreover, we can represent the graph with sequences of its ordered nodes or edges denoted by $S^{node,\ \pi}$ and $S^{edge,\ \pi}$, respectively, where the former is denoted by the shorter form $S^\pi$ in the following for simplicity. The notations are summarized in Table \ref{tab:notations}.
\subsection{Problem Formulation}
Given a dataset of graphs $\mathcal{D}_G$ with the underlying data distribution $p(G)$ (i.e., for each graph $G$ in the dataset, $G \sim p(G)$), a DGG aims to learn how to obtain new samples from the data distribution by employing deep neural networks. Specifically, this can be done by either estimating the  $p(G)$ first and then sample from the estimated distribution or acquiring an implicit strategy, which only learns how to sample from the distribution without explicitly modeling it.
\section{Autoregressive Deep Graph Generators}\label{sec:autoregressive}
In this section, we review those approaches generating graph structures sequentially in a step-wise fashion, where the prediction at each time step is affected by the previous outputs. We further divide them into recurrent and non-recurrent approaches, where the former captures the generation history by employing recurrent units, while the latter makes decisions directly based on the latest partially generated graph. The main characteristics of these methods are summarized in Table \ref{tab:autoregressive}.
\subsection{Recurrent DGGs}
Recurrent DGGs are a bunch of autoregressive deep graph generators that use RNNs, namely long short-term memory (LSTM) \cite{hochreiter1997long} or gated recurrent units (GRU)\cite{cho2014learning}, to exert the influence of the generation history on the current decision. Here, we provide a detailed review of these methods in two subcategories.
\subsubsection{Node-by-Node Generators}
Most of the autoregressive methods append one new node at a time into the already generated graph. For example, Li et al. \cite{li2018multi} propose to generate molecular graphs sequentially, where the generation process initiates by adding a node to an empty graph. It then continues by iteratively deciding whether to append a new node to the graph, connect the lastly added node to the previous ones, or terminate the process. To this end, the authors propose two architectures, namely MolMP and MolRNN, to determine probabilities for each of these three actions. More precisely, MolMP decides based on the graph's current state, modeling the generation as a Markov Decision Process. It first calculates an initial embedding for graph nodes followed by several convolutional layers and an aggregation operation to obtain a graph-level representation. It then passes both the node-level and graph-level embeddings through MLP and softmax layers to compute the probabilities required for action selection. MolRNN, on the other hand, exploits molecule level recurrent units to make the generation history affect the current decision, which improves the model's performance. It adopts the same approach as MolMP to obtain embeddings and then updates the recurrent units' hidden state as follows:
\begin{equation}
h_i = f_{trans}(h_{i-1}, h_{v^*}, h_{G_{i-1}}),
\end{equation}
where $f_{trans}$ is implemented using GRUs, $h_{v^*}$ is the latest appended node embedding, and $h_{G_{i-1}}$ denotes the representation for the graph generated before the $i$-th generation step. Next, the action probabilities are calculated similarly as MolMP, except that MolRNN replaces $h_{i}$ by the graph-level representation. Moreover, the authors make the conditional graph generation possible by first converting a given requirement to a conditional code and then modifying the graph convolution to include this code.\par
\begin{figure}[t]
\centering
\includegraphics[width=1\textwidth]{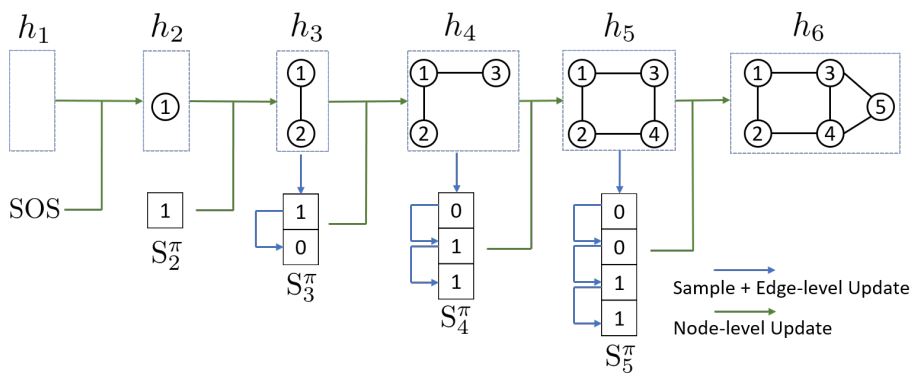}
\vspace*{-1.3cm}
\begin{center}
\caption{An illustration of the graph generation procedure at inference time proposed
in GraphRNN \cite{you2018graphrnn} (reprinted with permission). Green arrows denote the
graph-level RNN, and blue arrows represent the edge-level RNN.}
\label{fig:GraphRNN}
\end{center}
\end{figure}

You et al. \cite{you2018graphrnn} propose GraphRNN, another deep autoregressive model with a hierarchical architecture consisting of a graph-level RNN and an edge-level RNN which learns to sample $G \sim p(G)$ without explicitly computing $p(G)$. For this purpose, GraphRNN first defines a mapping $f_S$ from graphs to sequences where for a graph $G$ with $n$ nodes under the node ordering $\pi$, the mapping is defined as follows:
\begin{equation}
S^\pi = f_S(G, \pi) = (S_1^\pi, ..., S_n^\pi),
\end{equation}
where each element $S_i^\pi \in \{0, 1\}^{i-1}, i\in\{1, ..., n\}$ represents the edges between node $\pi(v_i)$
and the previous nodes. Since for undirected graphs, there exists the mapping function $f_G(S^\pi) = G$, it is possible to sample $G$ at inference time by first sampling $S^\pi \sim p(S^\pi)$ and then applying $f_G$, which obviates the need to compute $p(G)$ explicitly. To learn $p(S^\pi)$, due to the sequential nature of $S^\pi$,
$p(S^\pi)$ can be further decomposed as in Eq. (\ref{eq:2}), which is modeled by an RNN with state transition and output functions defined in Eq. (\ref{eq:3}) and (\ref{eq:4}), respectively:
\begin{equation}\label{eq:2}
p(S^\pi) =\prod_{i=1}^{n+1}p(S_i^\pi |S_1^\pi, ..., S_{i-1}^\pi)=\prod_{i=1}^{n+1}p(S_i^\pi |S_{<i}^\pi),
\end{equation}
\begin{equation}\label{eq:3}
h^{node}_i = f_{trans}^{node}(h^{node}_{i-1}, \text{In}^{node}_{i}), \ \ \text{In}^{node}_{i} = S_{i-1}^\pi,
\end{equation}
\begin{equation}\label{eq:4}
\theta_i = f_{out}(h_i^{node}),
\end{equation}
where $f_{trans}^{node}$ is implemented using a GRU and serves as the graph-level RNN that maintains the state of the graph generated so far. Furthermore, the authors propose two varients for the implementation of $f_{out}$. First, they propose GraphRNN-S, a simple variant that does not consider dependencies between edges and models $p(S_i^\pi |S_{<i}^\pi)$ as a multivariate Bernoulli distribution. Next, to fully capture the complex edge dependencies, they propose the full GraphRNN model as illustrated in Figure \ref{fig:GraphRNN}, which approximates $f_{out}$ by another RNN (i.e., the edge-level RNN) formulated as follows:
\begin{equation}\label{eq:7}
\small
h_{i, j}^{edge} = f_{trans}^{edge}(h^{edge}_{i, j-1}, \text{In}^{edge}_{j}), \ \ \text{In}^{edge}_{j} = S_{i, j-1}^\pi, \ \ h^{edge}_{i,0} = h^{node}_i.
\end{equation}
Furthermore, GraphRNN introduces a BFS node ordering scheme to improve scalability with two benefits. First,  it will suffice to train the model on all possible BFS orderings, rather than all possible node permutations. Second, it reduces the number of edges to be predicted in the edge-level RNN. 
\par
Subsequently, several graph generation methods have been proposed inspired by GraphRNN. For example, Liu et al. \cite{liuauto2019} propose two further variants for the implementation of $f_{out}$ function in Eq. (\ref{eq:4}), namely RNN-Transf and GraphRNN+Attn. Specifically, RNN-Transf replaces the edge-level RNN in the full GraphRNN model with a vanilla Transformer \cite{vaswani2017attention} decoder consisting of a self-attention sublayer and a graph-state attention sublayer with the memory from hidden states of the node-level RNN. GraphRNN+Attn, on the other hand, maintains the edge-level RNN. More precisely, it is an additive attention mechanism in the edge-level RNN that computes the attention weights in each step using the last hidden states of the node-level RNN as well as the current hidden state of the edge-level RNN.\par
MolecularRNN\cite{popova2019molecularrnn} extends GraphRNN to generate realistic molecular graphs with desired chemical properties. As in molecular graphs, both nodes and edges have types, likelihood formulation in Equation (\ref{eq:2}) is rewritten as follows:
\begin{equation}\label{eq:5}
p(S^\pi, C^\pi) =\prod_{i=1}^{n+1} p(C_i^\pi|S_{<i}^\pi , C_{<i}^\pi)p(S_i^\pi|C_i^\pi, S_{<i}^\pi,  C_{<i}^\pi ),
\end{equation}
where $S_{i,j}^\pi \in \{0, 1, 2, 3\}$ is the categorical edge type that corresponds to no, single, double, or triple bonds, and $C_{i}^\pi \in \{1, 2, ..., K\}$ determines node (atom) type. Then, MolecularRNN substitutes the graph-level RNN input in Eq. (\ref{eq:3}) with the embeddings of categorical inputs as in Eq. (\ref{eq:6}):
\begin{equation}\label{eq:6}
\text{In}_{i}^{node} = [emb(S_{i-1}^\pi),emb(C_{i-1}^\pi)].
\end{equation}
Furthermore, a two-layer MLP with softmax output activation is added on top of the hidden states of both graph-level and edge-level RNNs to predict node and edge types, respectively.
After likelihood pretraining on the molecular datasets, the model is fine-tuned with the policy gradient algorithm to shift the distribution
of the generated samples to some desired chemical properties, namely, lipophilicity, drug-likeness, and melting point. Thus, the MolecularRNN acts as a policy network to output the probability of the next action given the current state, where the set of states consists of all possible sub-graphs and the possible atom connections to the existing graph, for all the atom types, serve as the action set. Moreover, each valid molecule is considered as a final state $s_n$, where its corresponding final reward is denoted by $r(s_n)$. The intermediate rewards $r(s_i), 0 < i < n$ are also obtained by discounting $r(s_n)$ as in the following loss function formula:
\begin{equation}\label{eq:8}
\mathcal{L}(\theta) = -\sum_{i=1}^n r(s_n). \gamma^i . \log p (s_i|s_{i-1}; \theta),
\end{equation}
where $\gamma$ is the discount factor and the transition probabilities $p(s_i|s_{i-1}; \theta)$ are the elements of the product in Eq(\ref{eq:5}). Furthermore, MolecularRNN introduces the structural penalty for atoms violating valency constraints during training. It also adopts a valency-based rejection sampling method during inference, which guarantees the generated samples' validity.

\par
Sun et al. \cite{sun2019graph} learn a mapping from a source to a target graph by adopting an encoder-decoder based approach, where the encoder utilizes recurrent based models to encode the source graph and the decoder generates the target graph in a node-by-node fashion, which makes it necessary to consider an ordering over nodes. Therefore, the authors first introduce a procedure to transform a graph $G$ into a DAG (Directed Acyclic Graph) to provide the required node ordering. They then obtain embeddings for each of the DAG's nodes by proposing two encoders: an Energy-Flow encoder and a Topology-Flow encoder, where the former utilizes only the information of adjacent nodes, while the latter exploits both the adjacent and non-adjacent nodes' information. Afterward, the decoder sequentially generates the target graph conditioned on the source graph by adopting a relatively similar generation strategy as the GraphRNN.\par
So far, we have studied GraphRNN \cite{you2018graphrnn} as one of the most widely used deep graph generators and then reviewed the subsequent graph generation approaches inspired by it; each generates different types of graphs from general to molecular ones. Moreover, there are also other methods that use GraphRNN as a basis for solving some application-specific problems. For example, REIN \cite{daroya2020rein} proposes to autoregressively generate meshes from input point clouds inspired by GraphRNN so that in each generation step, it predicts edges from the newly introduced point to all the previous ones. The generated mesh can then be used for the task of 3D object reconstruction. DeepNC \cite{tran2019deepnc} is another GraphRNN-based approach that proposes a network completion algorithm to infer the missing parts of a network. Specifically, it first trains GraphRNN to learn a likelihood over the data distribution. The method then formulates an optimization problem to infer the missing parts of a partially observed input graph in such a way that maximizes the learned likelihood.
\subsubsection{Edge-by-Edge Generators}
In addition to the methods discussed so far, there also exist other approaches adopting an edge-based generation strategy.  Bacciu et al. \cite{bacciu2020edge} propose to generate a sequence of edges for each graph instead of generating graphs node-by-node. They first convert a graph $G$ under the node ordering $\pi$ to an ordered edge sequence $S^{edge,\ \pi} = [S^{edge,\ \pi}_1, ..., S^{edge,\ \pi}_m]$, where $S^{edge,\ \pi}_i = (u_i^\pi, v_i^\pi)$ is the $i$-th edge in the sequence that connects the source node $u_i^\pi$ to the destination node $v_i^\pi$ (here $u_i^\pi$ and $v_i^\pi$ are IDs assigend to graph nodes by $\pi$). Note that the sequence is ordered, that is $S_i^{edge,\ \pi} \leq S_{i+1}^{edge,\ \pi}$ iff $u_i^\pi < u_{i+1}^\pi$ or ($u_i^\pi = u_{i+1}^\pi$ and $v_i^\pi < v_{i+1}^\pi$). Then, the authors define $U^\pi = [u_1^\pi, . . . , u_m^\pi]$ and $V^\pi = [v_1^\pi, . . . , v_m^\pi]$ as sequences of the source and destination node IDs, respectively and decompose the edge sequence probability as followes:
\begin{equation}
S^{edge,\ \pi} = p(U^\pi)p(V^\pi|U^\pi),
\end{equation}
where $p(U^\pi)$ and $p(V^\pi|U^\pi)$ are approximated with two RNNs. Specifically, \texttt{RNN1} is used to estimate $p(U^\pi)$ with the following transition and output functions:
\begin{equation}
\begin{split}
&h_i^\texttt{RNN1} = f_{trans}(h_{i-1}^\texttt{RNN1}, \text{In}_{i}^\texttt{RNN1}), \ \ \ \text{In}_{i}^\texttt{RNN1} = emb(u_{i-1}^\pi)
\\
&p(u_i^\pi|u_{i-1}^\pi, h_{i-1}^\texttt{RNN1}) = f_{out}(h_i^\texttt{RNN1})=\text{Softmax}(Lin(h_i^\texttt{RNN1})),
\end{split}
\end{equation}
where $f_{trans}$ is implemented as a GRU and $Lin$ is a linear projection to map the recurrent output to the node ID space. Once all of the $U^\pi$'s elements are generated, the last recurrent state of \texttt{RNN1} is used to initialize the state of \texttt{RNN2}, and the process moves to \texttt{RNN2} that is given $U^\pi$ as input. Thus \texttt{RNN2} computes the probability distribution of $p(V^\pi|U^\pi)$ with the same architecture as \texttt{RNN1} by approximating $p(v_i | u_i, h_{i-1}^\texttt{RNN2})$ each step. 
\par
\begin{figure}[t]
\centering
\includegraphics[width=1\textwidth]{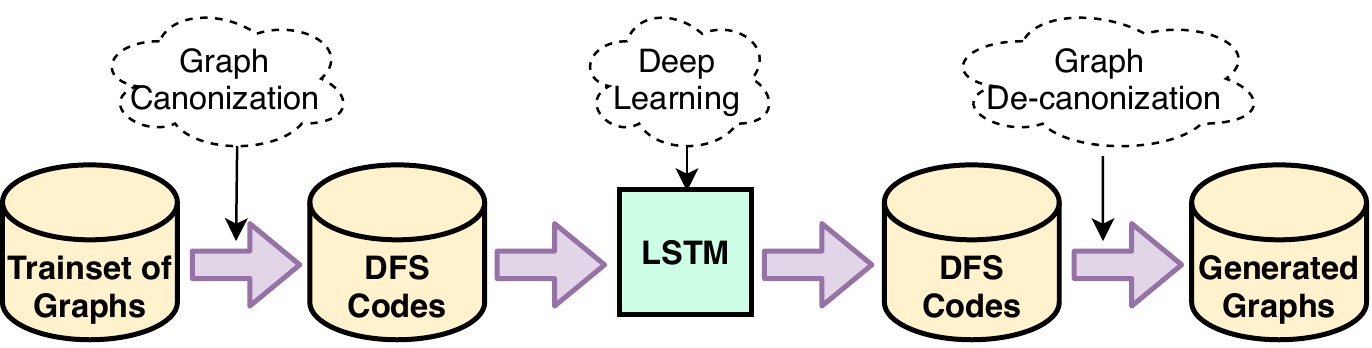}
\vspace*{-1cm}
\begin{center}
\caption{Flowchart of GraphGen \cite{goyal2020graphgen}.}
\label{fig:GraphGen}
\end{center}
\end{figure}

Similarly, GraphGen \cite{goyal2020graphgen} proposes another edge-based generation strategy that adds a single edge to the already generated graph at each stage. To this end, the method first converts a graph $G$ to a sequence $ S^{edge} = [S_1^{edge},...,S_m^{edge}]$ using the minimum DFS code\cite{yan2002gspan}, where each $S_i^{edge}$ corresponds to an edge $e = (u,v)$ and is described using a 5-tuple $(t_u, t_v , L_u, L_e, L_v)$, where $t_u$ is the timestamp assigned to node $u$ during the DFS traversal, and $L_u$ and $L_e$ denote the node and edge labels, respectively. As the minimum DFS codes are canonical labels, and thus there is a one-to-one mapping between a graph and its corresponding sequence,  there is no longer need to deal with multiple representations for the same graph under different node permutations during training, which improves the method scalability. Then, GraphGen takes a similar approach to GraphRNN \cite{you2018graphrnn} to decompose $p(S^{edge})$ as follows:
\begin{equation}\label{GraphGen-likelihood-eq}
\small
p(S^{edge}) =\prod_{i=1}^{m+1}p(S_i^{edge} |S_1^{edge}, ..., S_{i-1}^{edge})=\prod_{i=1}^{m+1}p(S_i^{edge} |S_{<i}^{edge}),
\end{equation}
where $m$ is the number of edges, and making the simplifying assumption that $t_u$, $t_v$, $L_u$, $L_e$, and $L_v$ are independent, reduces  $p(S_i^{edge} |S_{<i}^{edge} )$ in Eq. (\ref{GraphGen-likelihood-eq}) to:
\begin{equation}\label{GraphGen-indep}
\begin{split}
p(S_i^{edge} |S_{<i}^{edge} ) &= p((t_u,t_v , L_u, L_e, L_v ) | S_{<i}^{edge} )\\&
= p(t_u |S_{<i}^{edge} ) \times p(t_v |S_{<i}^{edge} ) \times p(L_u |S_{<i}^{edge} )\\&
\times p(L_e |S_{<i}^{edge} ) \times p(L_v |S_{<i}^{edge} ).
\end{split}
\end{equation}
To capture conditional distributions in Eq. (\ref{GraphGen-indep}), the authors propose to use a custom LSTM with the transition function $f_{trans}$ in Eq. (\ref{GraphGen-lstm}), and five separate output functions for each component of the 5-tuple. For example, $f_{t_u}$ in Eq. (\ref{GraphGen-out-prediction}) is utilized for predicting $t_u$, where $\sim_M$ represents sampling from a multinomial distribution. 
\begin{equation}\label{GraphGen-lstm}
h_i = f_{trans} (h_{i-1}, \text{In}_{i}), \ \ \text{In}_{i} = emb (S_{i-1}^{edge}),
\end{equation}
\begin{equation}\label{GraphGen-out-prediction}
t_u \sim_M\ \theta_{t_u} = f_{t_u} (h_i ),
\end{equation}
\begin{equation}
S_i^{edge} = concat(t_u, t_v, L_u, L_e, L_v).
\end{equation}
Figure \ref{fig:GraphGen} outlines the proposed pipeline.
\subsection{Non-Recurrent DGGs}
There are some other autoregressive methods that, unlike the recurrent models, do not consider the entire history, and instead, they only focus on the latest version of the partially generated graph at each time step. To better review these methods, we further divide them into two following subsections.
\subsubsection{Attention-Based Methods}
\begin{figure}[t]
\centering
\includegraphics[width=1.03\textwidth]{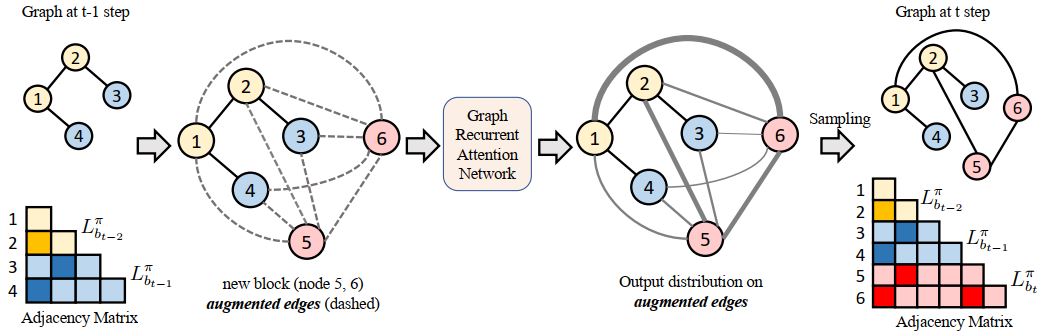}
\vspace*{-1.3cm}
\begin{center}
\caption{An overview of GRAN \cite{liao2019efficient} (reprinted with permission). Dashed lines are augmented edges. Nodes with the same color belong to the same block (block size = 2). }
\label{GRAN-fig}
\end{center}
\end{figure}

Here, we review the methods in which the attention mechanism plays a key role. In this regard, GRAN \cite{liao2019efficient} proposes to generate one block of nodes and associated edges at each generation step by optimizing the following likelihood:
\begin{equation}\label{GRAN-likelihood}
p(L^\pi)=\prod_{t=1}^Tp(L^\pi_{\textbf{b}_{\textbf{t}}}|L^\pi_{\textbf{b}_{\textbf{1}}},..., L^\pi_{\textbf{b}_{\textbf{t-1}}}),
\end{equation}
where $L^\pi$ is the lower triangular part of the adjacency matrix $A^\pi$, $B$ denotes the block size, $\textbf{b}_{\textbf{t}} = \{B(t - 1) + 1,..., Bt\}$ is the set of row indices for the $t$-th block of $L^\pi$, and $T = \ceil{\frac{N}{B}}$ is the number of graph generation steps. For the $t$-th step, GRAN adds $B$ new nodes to the already-generated subgraph and connects them with each other as well as the previous $B(t - 1)$ nodes to acquire an augmented graph as depicted in Figure \ref{GRAN-fig}. The authors then apply the following graph neural network with attentive messages on the augmented graph to get updated node representations:
\begin{equation}\label{GRAN-GNN}
\begin{split}
m^r_{ij} = f(h^r_i - h^r_j)\textbf{,} \ \ \ \tilde{h^r_i}&= [h^r_i, x_i]\textbf{,}  \ \ \ a^r_{ij} = \sigma\big( g(\tilde h^r_i - \tilde h^r_j)\big)\\
h^{r+1}_i &= \text{GRU}(h^r_i, \sum_{j\in \mathcal{N}(i)}a^r_{ij}m^r_{ij}),
\end{split}
\end{equation}
where $h^r_i$ is the representation for node $i$ after round $r$, $m^r_{ij}$ is the message vector from node $i$ to $j$, $x_i$ indicates whether node $i$ is in the previously generated nodes or the newly added ones, and $a^r_{ij}$ is an attention weight associated with $edge (i, j)$. Both the message function $f$ and the attention function $g$ are implemented as 2-layer MLPs with ReLU nonlinearities. After $R$ rounds of message passing, the final node representation vectors $h^R_i$ for each node $i$ is obtained, and then GRAN models the conditional probability in Eq. (\ref{GRAN-likelihood}) with a mixture of Bernoulli distributions to capture edge dependencies via $K$ latent mixture components:
\begin{equation}\label{GRAN-Mixture}
\begin{split}
&p(L^\pi_{\textbf{b}_{\textbf{t}}}|L^\pi_{\textbf{b}_{\textbf{1}}},..., L^\pi_{\textbf{b}_{\textbf{t-1}}}) = \sum_{k=1}^K\alpha_k\prod_{i\in \textbf{b}_{\textbf{t}}}\prod_{1\leq j\leq i} \theta_{k, i, j}
\\
&\alpha_1, ..., \alpha_K = \text{Softmax}\ \Big(\sum_{i\in \textbf{b}_{\textbf{t}}, 1\leq j\leq i}\text{MLP}_{\alpha}(h_i^R-h_j^R) \Big)
\\
&\theta_{1, i, j}, ...,  \theta_{K,i,j} = \sigma \big(\text{MLP}_\theta(h^R_i - h^R_j)\big).
\end{split}
\end{equation}
\par
GRAM \cite{kawai2019scalable} proposes to combine graph convolutional networks with graph attention mechanisms to obtain richer features during the graph generation process, where the proposed graph attention mechanism extends the one in \cite{vaswani2017attention} by introducing bias terms as a function of the shortest path between nodes. In particular, GRAM tries to maximize the following likelihood, which is somewhat similar to Eq. (\ref{eq:5}):
\begin{equation}
\small
\begin{split}
p(&A^\pi, C^\pi) =\\ &\prod_{i=1}^{n+1} p(C_i^\pi|A_{<i, <i}^\pi, C_{<i}^\pi) \prod_{j=1}^{i-1}p(A_{j, i}^\pi|A_{<j,i}^\pi, C_i^\pi, A_{<i,<i}^\pi, C_{<i}^\pi).
\end{split}
\end{equation}
To this end, the authors propose an architecture that consists of three networks, namely, feature extractor, node estimator, and edge estimator. Firstly, the feature extractor extracts the local and global information using graph convolution layers and graph attention layers, respectively, where an attention layer employs a self-attention mechanism with the query, key, and value that are set to the node feature vectors. A graph pooling layer then aggregates all node features into a graph feature vector, denoted as $h^G$, by summing them up. Next, the node estimator determines a label for the new node based on the feature vector of the graph generated so far. Thereafter, the edge estimator predicts labels for edges between the newly added node and those already exist in the graph one after the other using a source-target attention mechanism as follows:
\begin{equation}
A_{j, i}^\pi = \text{Softmax}(g_{EE}(h^v_j,  h^G, h^v_i, h^e_{<j})),
\end{equation}
where $g_{EE}$ is a three-layer feedforward network, $h^v_j$ and $h^v_i$ are label embeddings of node $v_j$ and the new node $v_i$, respectively, and $h^e_{<j}$ is computed using a source-target attention with $\text{Concat}(h^v_j, h^v_i)$ as its query and $\{\text{Concat}(h^v_t, h^v_i, h^e_{t,i})|t = 1, ..., j - 1\}$ as both the key and value. \par
AGE \cite{fan2020attention} introduces another attention-based generative model, which is conditioned on some input graphs. In other words, the method takes an existing source graph as input and generates a transformed version of it, modeling its evolution. To this end, the authors propose an encoder-decoder based architecture, where its decoder autoregressively generates the target graph in a node-by-node fashion. More specifically, the encoder first applies the self-attention mechanism to the source graph in order to learn its nodes' representations. Then, at each generation step, the decoder first adopts a similar self-attention mechanism as the encoder, followed by source-target attention, which discovers the correlations between the nodes in the source graph and the ones in the already generated target graph. This way the decoder computes a representation for the graph generated so far, which will be further used to predict the new node's label and connections.
\subsubsection{Other Methods}
\begin{figure*}[tp]
\centering
\includegraphics[width=0.65\textwidth]{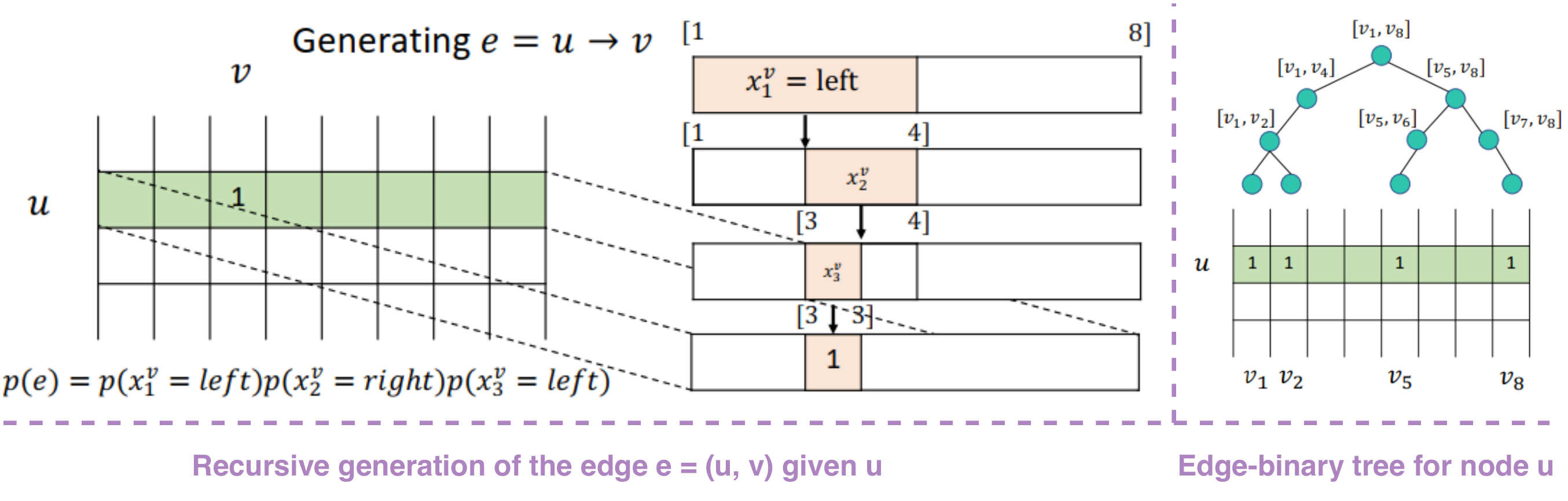}
\vspace*{-0.2cm}
\centering\caption{An overview of the edge generation procedure in \cite{dai2020scalable} (reprinted with permission).}
\label{BiGG-fig}
\end{figure*}
\begin{table*}[ht!]
\begin{center}
\small
 \begin{tabular}{||l l l l l l||}
 \hline
 Method & Recurrent & \makecell[l]{Generation\\ Strategy} & \makecell[l]{Attention\\ Mechanism} & Features & \makecell[l]{Conditional\\ Generation}\\ [0.5ex] 
 \hline\hline
 MolMP\cite{li2018multi} & No & Node-by-node& No& Node/Edge& Yes \\ 
 \hline
 MolRNN\cite{li2018multi} & Yes & Node-by-node& No& Node/Edge& Yes \\ 
 \hline
 GraphRNN\cite{you2018graphrnn} & Yes & Node-by-node& No& -& No \\ 
 \hline
 MolecularRNN\cite{popova2019molecularrnn} & Yes & Node-by-node& No& Node/Edge& No \\
 \hline
Bacciu et al. \cite{bacciu2019graph, bacciu2020edge} & Yes & Edge-by-edge& No& -& No \\
 \hline
GraphGen \cite{goyal2020graphgen} & Yes & Edge-by-edge& No& Node/Edge& No \\
\hline
GRAN\cite{liao2019efficient} & No & Block of nodes& Yes& -& No \\
 \hline
GRAM \cite{kawai2019scalable} & No & Node-by-node& Yes& Node/Edge& No \\
\hline
AGE \cite{fan2020attention} & No & Node-by-node & Yes& Node& Yes\\ 
\hline
DeepGMG\cite{li2018learning} & No & Node-by-node & Yes& Node/Edge& Yes\\ 
\hline
BiGG \cite{dai2020scalable} & No & Node-by-node & No& -& No\\ 
[1ex] 
 \hline
\end{tabular}
\caption{\label{tab:autoregressive} The Main Characteristics of Autoregressive Deep Graph Generators}
\end{center}

\end{table*}
Besides the attention-based methods reviewed above, other autoregressive non-recurrent DGGs either do not use attention at all, or the attention mechanism does not play a decisive role in their generation process. For example, DeepGMG \cite{li2018learning} proposes a sequential graph generation process which can be seen as the following sequence of decisions: (1) whether to add a new node of a particular type or not (with probabilities provided by the $f_{addnode}$ in Eq. (\ref{DeepGMG-addnode}), where $h_G$ is the graph representation vector, and $f_{an}$ is an MLP that maps $h_G$ to the action output space), if a node type is selected (2) the model decides whether to continue connecting the newly added node to the existing graph or not (referring to Eq. (\ref{DeepGMG-addedge}), where $h_v^{(T)}$ is embedding of the new node $v$ after $T$ rounds of propagation in a graph neural
network, and $f_{ae}$ is another MLP), if yes (3) it selects a node already in the graph and connects it to the new node (referring to the Eq. (\ref{DeepGMG-nodes}), where $f_s$ maps pairs $h_u^{(T)}$ and $h_v^{(T)}$ to a score $s_u$). The algorithm goes back to step (2) and repeats until the model decides not to add another edge. Finally, the algorithm goes back to step (1) to add subsequent nodes or terminate the process.
\begin{equation}\label{DeepGMG-addnode}
f_{addnode}(G) = \text{Softmax}\ (f_{an}(h_G)),
\end{equation}
\begin{equation}\label{DeepGMG-addedge}
f_{addedge}(G, v) = \sigma(f_{ae}(h_G, h_v^{(T)})),
\end{equation}
\begin{equation}\label{DeepGMG-nodes}
\begin{split}
s_u = f_s(h_u^{(T)}, h_v^{(T)}),\ \  \forall u\in V\\
f_{nodes}(G, v) = \text{Softmax}\ (s).
\end{split}
\end{equation}
\par
DeepGG\cite{stier2020deep} further extends DeepGMG\cite{li2018learning} by adding the idea of finite state machines into the generation process. Furthermore, similar to the GraphRNN\cite{you2018graphrnn}, DeepGG learns the graph distribution from a sequence called construction sequence, which consists of graph evolutionary actions such as node addition, edge addition, and node deletion.\par
Recently, BiGG \cite{dai2020scalable} proposes an autoregressive model to increase scalability for generating sparse graphs. To learn a generative model, BiGG  uses a single canonical ordering $\pi(G)$ to model each graph G, as in \cite{li2018learning}, aiming to learn a lower bound on p(G):
\begin{equation}
\small
p(G) = p(V)P(E|V)=p(|V|=n)\sum_\pi p(A^\pi)\approx p(|V|=n) p(A^{\pi(G)}),
\end{equation}
where $p(|V|=n)$ can be directly estimated using an empirical distribution over the graph size. Therefore the goal is only to model $p(A^{\pi(G)})$ under a default canonical ordering, which will be denoted by $p(A)$ in the following. Considering that most real-world graphs are sparse, BiGG generates only the non-zero entries in $A$ in a row-wise manner to enhance efficiency and  scalability; thus, the method adopts a recursive strategy inspired by R-MAT \cite{chakrabarti2004r}, for generating each edge as illustrated in the left half of Figure \ref{BiGG-fig}. To further improve efficiency, the authors propose to jointly generate all the connections of an arbitrary node $u$ (non-zero entries in the $u$-th row of $A$) by autoregressively generating an {\it edge-binary tree}, as shown in the right half of Figure \ref{BiGG-fig}. Finally, BiGG introduces the full autoregressive model that generates the entire adjacency matrix row by row. The full model utilizes the autoregressive models as building blocks:
\begin{equation}
p(A) = p(\{\mathcal{N}_u\}_{u\in V})=\prod_{u\in V}p(\mathcal{N}_u|\{\mathcal{N}_{u'}:u'<u\}),
\end{equation}
where $\mathcal{N}_u$ denotes the set of neighbors for node $u$. More specifically, inspired by Fenwick tree \cite{fenwick1994new}, the authors propose a data structure called {\it row-binary forest} to encode all the {\it edge-binary trees} generated so far, which will be used to generate new {\it edge-binary tree} for the current step.
\section{Autoencoder-Based Deep Graph Generators} \label{sec:autoencoder}
This section reviews those approaches that employ whether autoencoders (AEs) or VAEs\cite{kingma2013auto} to generate graph structures. In particular, a common practice in these methods is first to encode an input graph into a latent space using GNN \cite{scarselli2008graph}, GCN \cite{kipf2016semi}, or their variants and then start to generate the graph from this latent space embedding. We divide the existing approaches based on their generation granularity level (i.e., adopting an all-at-once generation strategy, using valid substructures as building blocks, or generating graphs in a node-by-node fashion) into the following three subsections. The main characteristics of the most prominent autoencoder-based graph generators are presented in Table \ref{tab:autoencoder}.\par
\subsection{One-Shot Generators}
A series of autoencoder-based DGGs generate the entire graph all at once. VGAE\cite{kipf2016variational} proposes a graph generation model that primarily aims to perform unsupervised learning on graphs based on the variational autoencoder \cite{kingma2013auto}. Given a graph $G$ with adjacency matrix $A$ and node feature matrix $X$, VGAE infers the latent matrix \textbf{Z} by a two-layer GCN \cite{kipf2016semi}:
\begin{equation}\label{VGAE-1}
\small
q(\textbf{Z}|X,A)=\prod_{i=1}^nq(\textbf{z}_{i}|X,A),\ \text{with}\ \ q(\textbf{z}_{i}|X,A)=\mathcal{N}(\textbf{z}_{i}|\mu_i,\text{diag}(\sigma_{i}^2)),
\end{equation}
where $\mu=\text{GCN}_{\mu}(X,A)$ is the matrix of mean vectors $\mu_{i}$; similarly $\log\sigma = \text{GCN}_\sigma(X, A)$.  Then, the generative model is designed as a simple inner product of latent variables as follows:
\begin{equation}
\small
p(A|\textbf{Z})=\prod_{i=1}^n\prod_{j=1}^np(A_{ij}|\textbf{z}_{i},\textbf{z}_{j}),\ \text{with}\ \ p(A_{ij} = 1 |\textbf{z}_{i}, \textbf{z}_{j})=\sigma(\textbf{z}_{i}^{T}\textbf{z}_{j}),
\end{equation}
where $\sigma (.)$ is the logistic sigmoid function. The model parameters are then learned by optimizing the VAE objective. However, the authors also proposed a more straightforward, non-probabilistic, and autoencoder-based version of the method called GAE. The main limitation of VGAE is that it can only learn from a single input graph. GraphVAE \cite{simonovsky2018graphvae}, on the other hand, proposes another VAE-based generative model that learns from a dataset of graphs. The method first embeds the input graph into continuous representation \textbf{z} using a graph convolution network \cite{simonovsky2017dynamic} as the encoder $q_{\phi}(\textbf{z}|G)$, where the dimensionality of \textbf{z} is relatively small in order to learn a high-level compression of the input data. Then, the decoder outputs a probabilistic fully-connected graph with a fairly small predefined maximum size directly at once denoted by $\tilde{G}$. The whole GraphVAE model is trained by minimizing the upper bound on negative log-likelihood as follows:
\begin{equation}\label{GraphVAE}
\mathcal{L}_{\theta, \phi}(G)=\mathbb{E}_{q_\phi(\textbf{z}|G)}[-\log_{}p_{\theta}(G|\textbf{z})]+KL[q_{\phi}(\textbf{z}|G)||p(\textbf{z})],
\end{equation}
where $q_\phi(\textbf{z}|G)$ and $p_\theta(G|\textbf{z})$ are the encoder posterior and the decoder generative
distributions respectively, and $\phi$ and $\theta$ are parameters to be learned. Since no particular ordering of nodes is imposed in neither G nor $\tilde{G}$, the authors further adopt an approximate graph matching algorithm for aligning G with $\tilde{G}$ in order to compute the likelihood $p_\theta(G|\textbf{z})$ in Eq. (\ref{GraphVAE}). However, the growth of GPU memory requirements, number of parameters, and graph matching complexity for larger graph sizes limit the applicability of GraphVAE only to generate smaller graphs.\par
MPGVAE \cite{flam2020graph} further improves GraphVAE by building a message passing neural network (MPNN) into the encoder and decoder of a VAE, eliminating the need for complex graph matching algorithms. In particular, the method first encodes a molecular graph using a variant of MPNNs \cite{gilmer2017neural} combined with a graph attention \cite{velivckovic2017graph} to aggregate the information over each node's neighbors. The encoder then obtains a graph-level representation using the set2set model \cite{vinyals2015order} and uses this representation to parametrize the posterior distribution $q_\phi(\textbf{z}|G)$. Next, the decoder samples $\textbf{z}\sim q_\phi(\textbf{z}|G)$ and projects it to a high dimensional space consisting of several vectors and then passes these vectors through an RNN to compute initial states for the graph nodes. Afterward, it uses an identical MPNN as the encoder to obtain the final representation for each edge and node. The decoder then reconstructs the graph by predicting the atom types and bond types based on these final representations.\par
\begin{figure}[t]
\centering
\includegraphics[width=0.85\textwidth]{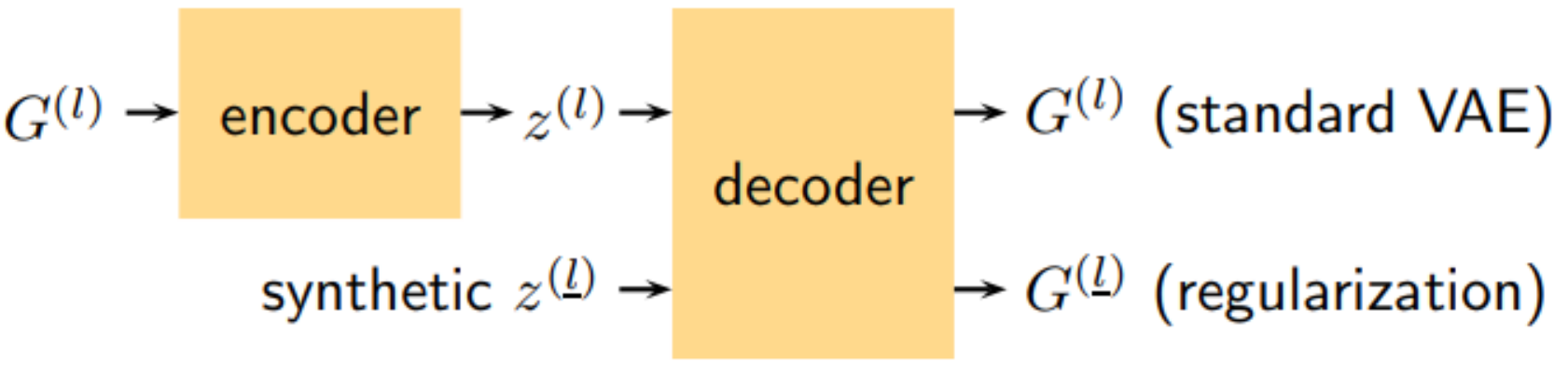}
\vspace*{-1cm}
\begin{center}
\caption{The framework of RGVAE \cite{ma2018constrained} (reprinted with permission). The top flow corresponds to the standard VAE, while the bottom flow denotes the regularization, where a synthetic $\textbf{z}^{(\underline{l})}$ is decoded to compute the constraints $g_i(\theta, \textbf{z}^{(\underline{l})})_+$.}
\label{fig:RGVAE}
\end{center}
\end{figure}

RGVAE \cite{ma2018constrained} regularizes the framework of variational autoencoders to generate semantically valid graphs. To impose validity constraints in the training of VAEs, RGVAE transforms a constrained optimization problem to a regularized, unconstrained one by adding inequality constraints to the objective function of VAEs, which forms a Lagrangian function. More precisely, RGVAE minimizes the following loss function in each parameter update:
\begin{equation}
\mathcal{L} = \mathcal{L}_{\theta, \phi}(G)+\mu \sum_ig_i(\theta, \textbf{z})_+, \ \ \ \ \text{where}\ \ \ \textbf{z}\sim p_{\theta}(\textbf{z}),
\end{equation}
where $g_i(\theta, \textbf{z}) \leq 0$ denotes the $i$-th validity constraint,  $g_+ = max(g, 0)$, and $\mathcal{L}_{\theta, \phi}(G)$ is the standard VAE loss function as in Eq. (\ref{GraphVAE}). The training of RGVAE is illustrated in Figure \ref{fig:RGVAE}, where $l$ denotes the index of a training example and $\underline{l}$ denotes a synthetic example, utilized in the regularization term.
\par
Graphite \cite{grover2019graphite} proposes a latent variable generative model based on VAE for unsupervised representation learning in large graphs. The method only models graph structure, and any supplementary information such as node features $X\in \mathbb{R}^{n\times k}$ is considered as conditioning evidence. To learn the model parameters $\theta$, Graphite maximizes a lower bound on log-likelihood of the observed adjacency matrix conditioned on $X$:
\begin{equation}
\log p_{\theta} (A|X)\geq \mathbb{E}_{q_\phi(\textbf{Z}|A, X)}\Big[\log \frac{p_\theta(A, \textbf{Z}|X)}{q_\phi(\textbf{Z}|A, X)}\Big].
\end{equation}
In more detail, the authors take an encoding approach based on the mean-field approximation, which represents graph nodes in the latent space using a graph neural network. Next, they propose an iterative two-step approach as the decoding part: it first constructs an intermediate weighted graph $\hat{A}$ from the latent matrix \textbf{Z}. Then, a parameterized graph neural network updates the latent matrix \textbf{Z}$^*$. The process alternates between these two steps to refine the graph gradually. More formally, given \textbf{Z} and $X$, Graphite iterates over the following two operations:
\begin{equation}
\hat{A}=\frac{\textbf{Z}\textbf{Z}^\top}{||\textbf{Z}||^2} +\textbf{11}^\top,\ \ \   \textbf{Z}^{*}=\text{GNN}_{\theta}(\hat{A},[\textbf{Z},X]),
\end{equation}
where an additional constant of \textbf{1} is added into the first operation to ensure entries are non-negative. Finally, it should be noted that similar to VGAE \cite{kipf2016variational}, Graphite is also limited to learning from a single input graph.\par
\begin{figure}[t]
\centering
\includegraphics[width=0.7\textwidth]{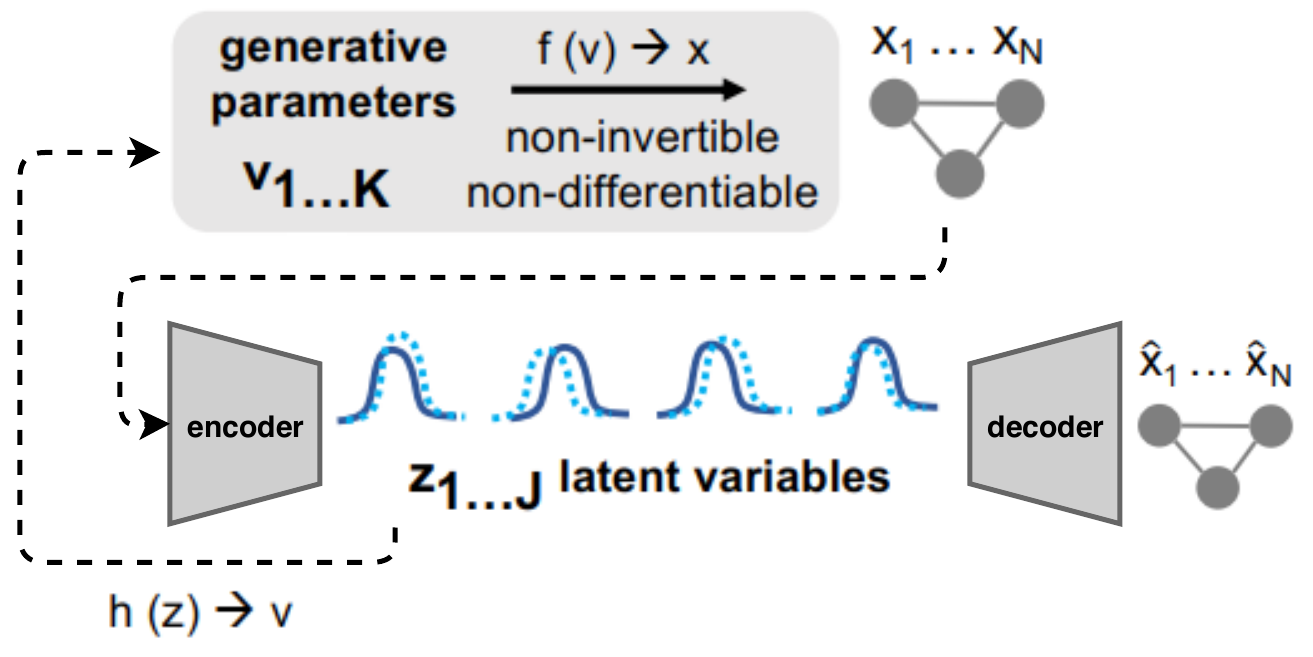}
\vspace*{-1cm}
\begin{center}
\caption{Architecture Overview of \cite{stoehr2019disentangling} (reprinted with permission).}
\label{fig:Disentangle}
\end{center}
\end{figure}
\begin{figure}[t]
\centering
\includegraphics[width=0.8\textwidth]{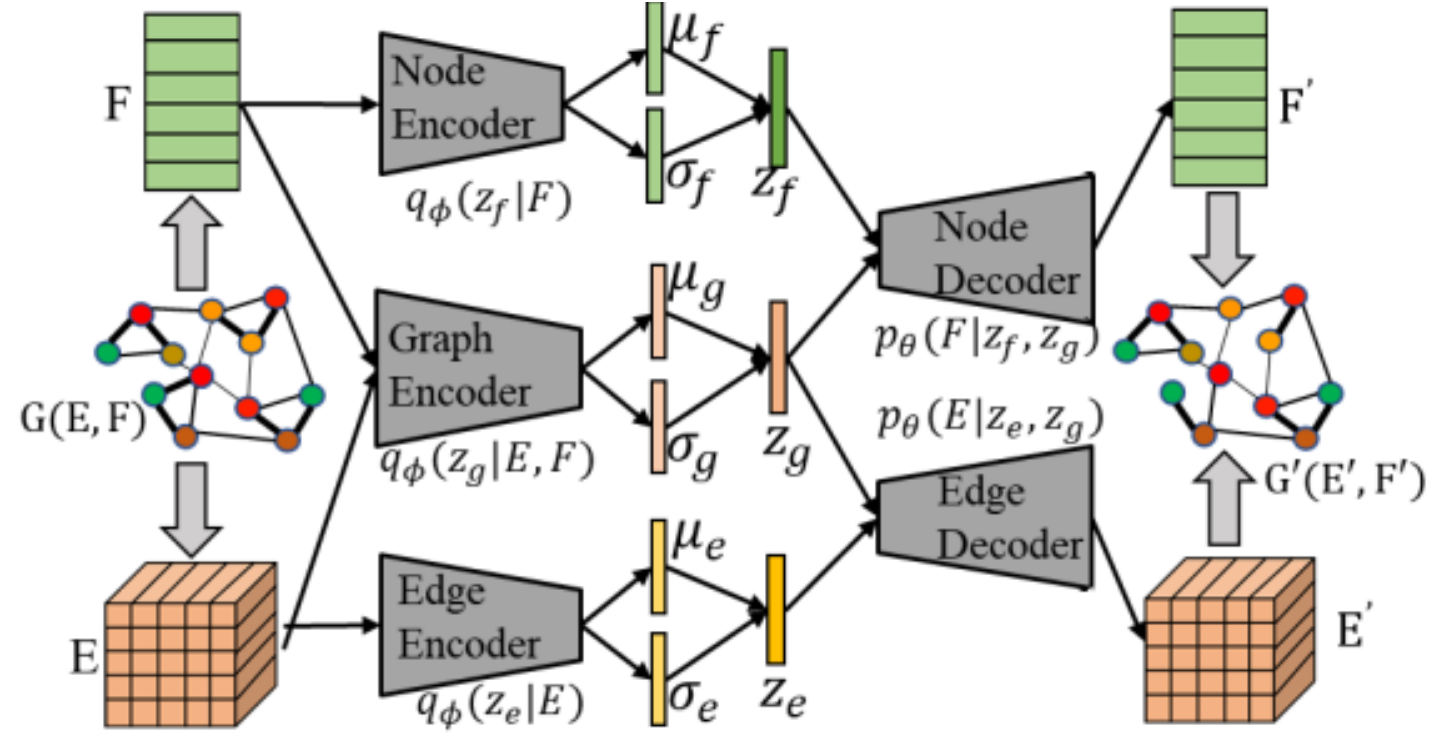}
\vspace*{-1cm}
\begin{center}
\caption{The architecture of NED-VAE \cite{guo2020interpretable} consisting of three sub-encoders, as well as two sub-decoders (reprinted with permission).}
\label{fig:NED-VAE}
\end{center}
\end{figure}

In addition to the aforementioned graph generative approaches, some initial steps have taken towards making DGGs interpretable. Stoehr et al. \cite{stoehr2019disentangling} propose to learn disentangled, interpretable latent variables corresponding to generative parameters of graphs. The main goal of learning such disentangled variables is to make the latent space more interpretable as each latent variable encodes one and only one data property. Therefore, the authors use a GCN as the encoder combined with a deconvolutional neural network as the decoder to minimize the loss function of $\beta$-VAE \cite{higgins2016beta}. In this setting, a higher value of $\beta$ yields the more orthogonalized latent space. To further enforce disentanglement of latent variables, the model also learns an additional \textit{parameter decoder} $h$, which maps latent variables to generative parameters as illustrated in Figure \ref{fig:Disentangle}. Recently, NED-VAE \cite{guo2020interpretable} proposes a more generalized generative approach for disentanglement learning on attributed graphs that uncovers the independent latent factors in both edges and nodes. In particular, NED-VAE aims to develop a model that can learn the joint distribution of the graph $G$ and three groups of generative independent latent variables, namely, $\textbf{z}_{f}, \textbf{z}_{e}$, and $\textbf{z}_{g}$  each of which controls the properties of only nodes, only edges, and the joint patterns between them, respectively. Therefore, inspired by the $\beta$-VAE \cite{higgins2016beta} formulation and considering the independence resulted from the disentanglement assumption, the goal is to maximize the following objective function:
\begin{equation}
\small
\begin{split}
   \mathcal{L}(\theta, \phi, G, \textbf{Z}, \beta)&=
\mathop{\mathbb{E}_{q_{\phi}(\textbf{Z}|G)}}[\log p_{\theta}(F|\textbf{z}_{f},\textbf{z}_{g})p_{\theta}(E|\textbf{z}_{e}, \textbf{z}_{g})]
    \\
   &-\beta D_{KL}(q_{\phi}(\textbf{z}_{f}|F)||p(\textbf{z}_{f}))
    -\beta D_{KL}(q_{\phi}(\textbf{z}_{e}|E)||p(\textbf{z}_{e}))\\
    &-\beta D_{KL}(q_{\phi}(\textbf{z}_{g}|E,F)||p(\textbf{z}_{g})),
\end{split}
\end{equation}
where $E \in \mathbb{R}^{n\times n\times d}$ is the edge attributes tensor, and $F \in \mathbb{R}^{n\times k}$ refers to the node attribute matrix. Based on the above objective, the authors propose an architecture consisting of three sub-encoders, namely, a node encoder, an edge encoder, and a node-edge co-encoder to model the distributions $q_{\phi}(\textbf{z}_{f}|F)$, $q_{\phi}(\textbf{z}_{e}|E)$, and $q_{\phi}(\textbf{z}_{g}|E,F)$, respectively, as depicted in Figure \ref{fig:NED-VAE}. The architecture also consists of two sub-decoders: a node decoder, and an edge decoder to model $p_{\theta}(F|\textbf{z}_{f},\textbf{z}_{g})$ and $p_{\theta}(E|\textbf{z}_{e},\textbf{z}_{g})$, respectively. The authors further propose multiple variant models to address different issues, including group-wise disentanglement, variable-wise disentanglement, and the trade-off between reconstruction error and disentanglement performance.

More recently, DGVAE \cite{li2020dirichlet} proposes to replace the commonly used Gaussian distribution in VAE-based graph generation models by the Dirichlet distribution as a prior for the latent variables, causing them to describe the graph cluster memberships, which as a result adds interpretability to the model. For this purpose, the authors adopt the same formulation as VGAE \cite{kipf2016variational} in Eq. (\ref{VGAE-1}) for the encoding process except that they utilize a GNN variant, named Heatts, proposed by their own and employ the Laplace approximation \cite{hennig2012kernel} to model both $q(\textbf{z}_{i}|X,A)$ and $p(\textbf{z}_{i})$ as Dirichlet distributions. Furthermore, DGVAE adopts a somewhat similar decoding strategy as VGAE \cite{kipf2016variational} and proves that maximizing the reconstruction term of the model is equivalent to minimizing \textit{balanced graph cut}, which further gives the authors the motivation for designing the Heatts.
\subsection{Substructure-Based Generators}
There exists a number of works using valid chemical substructures as building blocks to generate more plausible molecular graphs. JT-VAE \cite{jin2018junction} adopts such a strategy by extending the variational autoencoder framework, which, as a consequence, avoids invalidity of the intermediate subgraphs. Specifically, JT-VAE first decomposes a molecular graph G into a junction tree $\tau_G$ to make the graph cycle-free, where each node in the tree represents a substructure of the molecule. Then, both G and $\tau_G$ are encoded to latent representations $\textbf{z}_G$ and $\textbf{z}_\tau$, respectively, using different encoders. More precisely, $\textbf{z}_\tau$ encodes the junction tree without capturing the exact mutual connections between substructures, while $\textbf{z}_G$ encodes the graph to capture the fine-grained connectivities. Afterward, JT-VAE reconstructs the junction tree from its latent representation $\textbf{z}_\tau$ using a tree-structured decoder, where a tree is generated node-by-node, in a top-down manner. Next, the authors introduce a graph decoder to reproduce the molecular graph based on the underlying predicted junction tree. Since there are potentially many molecules corresponding to the same junction tree, the graph decoder learns how to assemble the subgraphs (nodes in the tree) to reconstruct the molecular graph. Furthermore, to generate molecules with desired properties, the method performs Bayesian optimization in the latent space.\par 
JT-VAE is basically designed to use small substructures as building blocks, which degrades its performance for generating larger molecules such as polymers. To address this issue, HierVAE \cite{jin2020hierarchical} recently proposes a motif-based hierarchical graph encoder-decoder that employs significantly larger motifs as basic building blocks. To this end, the authors design an encoder that learns hierarchical representation for a given molecular graph $G$ in a fine-to-coarse fashion, from atoms to connected motifs. Then, it obtains the latent vector  $\textbf{z}_G$ by sampling from a Gaussian distribution parametrized using the acquired motif representations. The decoder, on the other hand, autoregressively generates a molecular graph in a coarse-to-fine fashion conditioned on $\textbf{z}_G$. More specifically, at each generation step, the decoder first predicts the next motif to be attached to the already generated graph. Then, it predicts the attachment points in the new motif, i.e., what atoms belong to the intersection of the new motif and its neighbor motifs. Finally, the decoder decides how the new motif should be attached to the current graph based on its predicted attachment points. The model parameters are learned by minimizing the VAE loss function formulated in Eq. (\ref{GraphVAE}). Furthermore, the authors extend the architecture to graph-to-graph translation \cite{jin2018learning} in order to induce desired properties in the generated molecules.
\par 
MHG-VAE \cite{kajino2019molecular} guides VAE to always generate valid molecular graphs by proposing {\it molecular hypergraph grammar} (MHG),  a special case of {\it hyperedge replacement grammar} (HRG) \cite{drewes1997hyperedge} for generating {\it molecular hypergraphs}, to encode chemical constraints. In particular, the proposed encoder consists of three parts as follows:
\begin{equation}\label{eq:MHG-VAE}
Enc = Enc_N \circ Enc_G \circ Enc_H,
\end{equation}
where $Enc_H$ first encodes a molecular graph into a molecular hypergraph, and $Enc_G$ represents the molecular hypergraph as a parse tree by leveraging MHG. Then, $Enc_N$ encodes the previously generated parse tree into the latent continuous space using a seq2seq GVAE \cite{kusner2017grammar}. The decoder, on the other hand, acts as an inversion to the encoder and applies production rules, including those with chemical substructure terminals. Finally, using Bayesian optimization, MHG-VAE optimizes the latent continuous space (and its corresponding molecules) towards desired properties.
\par
\begin{figure}[t]
\centering
\includegraphics[width=0.7\textwidth]{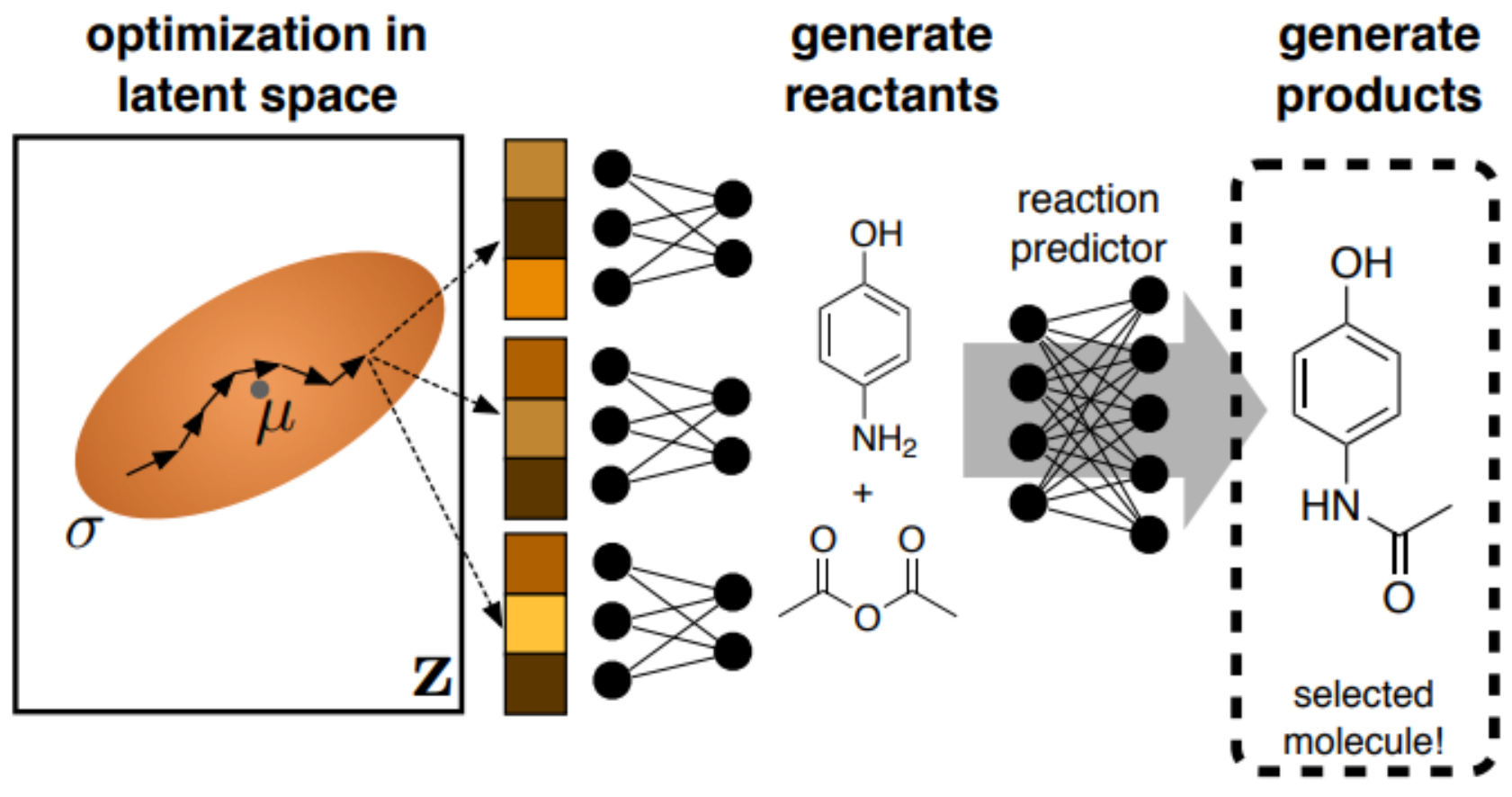}
\vspace*{-1cm}
\begin{center}
\caption{An overview of MoleculeChef \cite{bradshaw2019model} (reprinted with permission).}
\label{fig:MoleculeChef}
\end{center}
\end{figure}

MoleculeChef \cite{bradshaw2019model} proposes to generate molecular graphs using a set of common reactant molecules as building blocks to address the synthesizability issue. In particular, the encoder maps from a multiset of reactants to a distribution over latent space. This is done by using GGNNs \cite{li2015gated} to embed each reactant molecule separately, which are further summed to form one embedding for the whole multiset. A feed-forward network is then used to parameterize a Gaussian distribution over the latent space. The decoder, on the other hand, autoregressively maps from the latent space to a multiset of reactants using an RNN, where the latent vector \textbf{z} initializes its hidden layer, and at each generation step, it outputs one reactant or halts the process. Afterward, a reaction predictor \cite{schwaller2019molecular} predicts how the previously generated reactants produce a final molecule as illustrated in Figure \ref{fig:MoleculeChef}. To learn the model parameters, MoleculeChef minimizes the following  WAE \cite{tolstikhin2018wasserstein} objective function:
\begin{equation}\label{MoleculeChef}
\begin{split}
\mathcal{L}_{\theta, \phi}(G)&=\mathbb{E}_{G\sim \mathcal{D}}\mathbb{E}_{q_\phi(\textbf{z}|G)}[-\log_{}p_{\theta}(G|\textbf{z})]\\
&+\lambda D(\mathbb{E}_{G\sim \mathcal{D}}[q_{\phi}(\textbf{z}|G)], p(\textbf{z})),
\end{split}
\end{equation}
where $D$ is a divergence measure, namely the maximum mean discrepancy (MMD). Finally, the authors propose to optimize the molecular properties in the continuous latent space in a similar manner to CGVAE \cite{liu2018constrained}, which we will discuss in the following subsection. 
\subsection{Node-by-Node Generators}
\begin{figure*}[!t]
\centering
\includegraphics[width=0.7\textwidth]{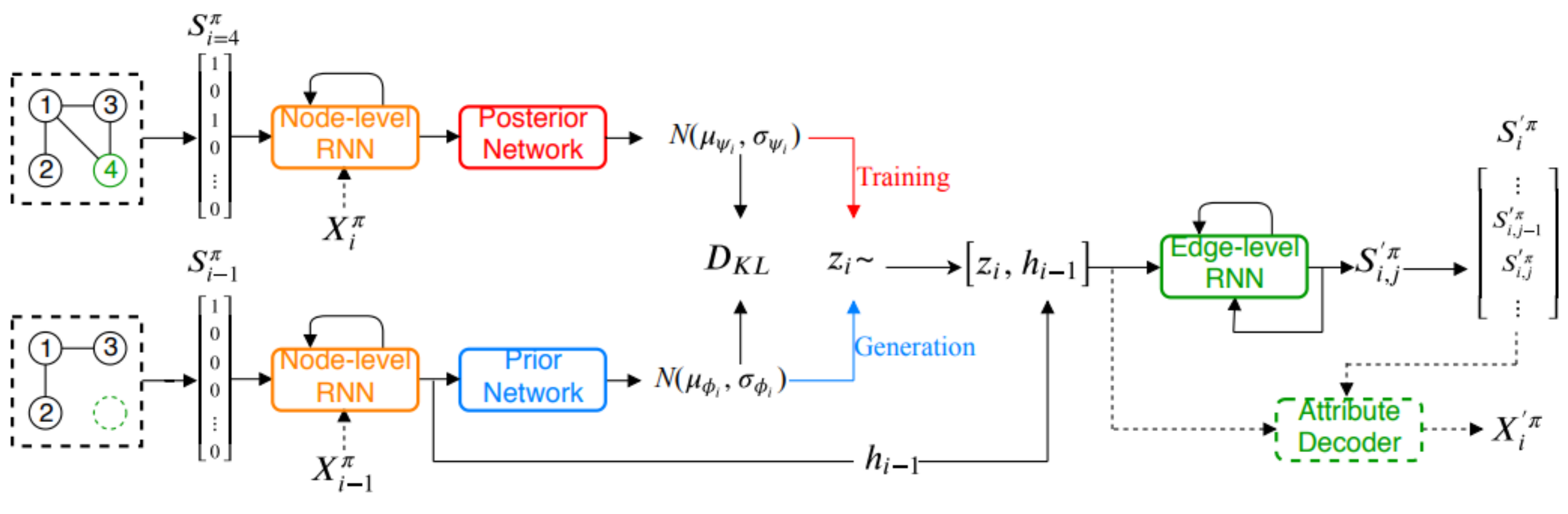}
\vspace*{-0.2cm}
\centering\caption{The framework of GraphVRNN \cite{su2019graph} (reprinted with permission).}
\label{GraphVRNN}
\end{figure*}
Besides the methods reviewed above, some other autoencoder-based DGGs use graph nodes as building blocks, and their decoders adopt autoregressive generation strategies. CGVAE \cite{liu2018constrained} is one of these methods whose encoder first samples a latent vector $\textbf{z}_v$ for each node $v$ of an input graph from a normal distribution parametrized using GGNNs\cite{li2015gated}. Then, the decoder starts from these vectors and sequentially generates a graph node-by-node with the help of two decision functions, namely, \texttt{focus} and \texttt{expand}. More precisely, the \texttt{focus} function determines the next node to be added into the graph, and the \texttt{expand} function iteratively chooses edges to add from the currently focused node until particular stop criteria met, and once the generated subgraph changes, all node representations get updated. Moreover, CGVAE employs a valency masking mechanism as part of \texttt{expand} function in the case of molecule generation to guarantee chemical validity. The authors also propose to optimize graph properties by minimizing $L_2$ distance between some numerical property Q and a differentiable gated regression score $R(G)$ as formulated in Eq. (\ref{CGVAE}), using gradient ascent in the continuous latent space:
\begin{equation}\label{CGVAE}
R(G) = \sum_v\sigma(g_1(\textbf{z}_v)).g_2(\textbf{z}_v),
\end{equation}
where $g_1$ and $g_2$ are neural networks. 
\par
DEFactor \cite{assouel2018defactor} proposes an encoder-decoder based architecture with a recurrent autoregressive decoder for conditional graph generation. In this regard, the encoder first applies a GCN \cite{kipf2016semi} to obtain node embeddings, which are then aggregated by an LSTM to compute the graph-level representation $\textbf{z}$. Then, the two-step decoder first employs another LSTM in order to autoregressively generate embeddings for each of the graph nodes based on the computed $\textbf{z}$:
\begin{equation}
h_i = f_{trans}(g_{in}([\textbf{z}, s_{i-1}]), h_{i-1}),\ \ 
s_i = f_{embed}([h_i, \textbf{z}]),
\end{equation}
where $f_{trans}$ is implemented by an LSTM \cite{hochreiter1997long}, $g_{in}$ and $f_{embed}$ are MLPs, and $s_i$ is the node embedding generated at timestep $i$. Next, the decoder establishes an edge factorization approach to compute the existence probability of an edge of type $k$ between nodes $u$ and $v$ as follows:
\begin{equation}
p(E_{u,v,k}|s_u, s_v)=\sigma(s_u^\intercal D_k s_v),
\end{equation}
where $D_k$ is the diagonal matrix of learnable factors for the $k$-th edge  type. Finally, DEFactor makes the generation process conditional by first concatenating the condition vector $C$ with $\textbf{z}$. It then utilizes a pre-trained discriminator to assess the property $C$ in the graphs generated by the decoder in the training phase.\par
NeVAE \cite{samanta2019nevae} proposes a probabilistic and permutation invariant encoder that is relatively similar to other graph representation learning algorithms, such as GraphSAGE \cite{hamilton2017inductive} and GCNs \cite{kipf2016semi}, except that it uses variational inference to learn the aggregator functions, which is further proved that makes the resulting embeddings well suited for the molecular graph generation task. The authors then introduce a probabilistic decoder that first samples the number of graph nodes from a Poisson distribution. It also samples a latent vector $\textbf{z}_v$ per node $v\in V$ from $\mathcal{N}(\textbf{0}, \textbf{I})$. Then, for each node $v$, the decoder passes $\textbf{z}_v$ through a neural network followed by a softmax classifier to determine node features, i.e., the atom type. Next, the total number of graph edges is sampled from a Poisson distribution parametrizes by another neural network conditioned on all latent vectors $\textbf{Z}$. Thereafter, the decoder samples graph edges one by one from a softmax distribution among all potential edges not generated that far, and similar to CGVAE \cite{liu2018constrained}, uses a set of binary masks to guarantee some local structural and functional properties. Finally, it determines the edge type by sampling from another softmax distribution with different binary masks. These masks get updated every time the decoder generates a new edge. The model's objective is similar to that of conventional VAE-based methods plus maximizing the Poisson distribution log-likelihood, which models the number of graph nodes. Moreover, similar to JT-VAE \cite{jin2018junction}, NeVAE utilizes Bayesian optimization over the continuous latent space to discover molecules with desirable properties.
\begin{table*}[ht!]
\begin{center}
\small
\resizebox{\columnwidth}{!}{%
 \begin{tabular}{||l l l l l l l||}
 \hline
 Method & Type & Input & \makecell[l]{Generation\\ Strategy} & \makecell[l]{Attention\\ Mechanism} & Features & 
\makecell[l]{Conditional\\ Generation}

\\ [0.5ex] 
 \hline\hline
  VGAE\cite{kipf2016variational} & AE/VAE & one single graph &All at Once& No & Node & No\\ 
 \hline
 GraphVAE \cite{simonovsky2018graphvae} & VAE & dataset of graphs & All at Once& No & Node/Edge & Yes\\ 
 \hline
 MPGVAE \cite{flam2020graph} & VAE & dataset of graphs & All at Once& Yes & Node/Edge & Yes\\ 
 \hline
 RGVAE \cite{ma2018constrained} & VAE & dataset of graphs & All at Once& No & Node/Edge & No\\
 \hline
 Graphite  \cite{grover2019graphite} & AE/VAE & one single graph & All at Once& No & Node & No\\
 \hline
 NED-VAE \cite{guo2020interpretable} & $\beta$-VAE & dataset of graphs & All at Once& No & Node/Edge & No\\ 
 \hline
 DGVAE \cite{li2020dirichlet} & VAE & one single graph & All at Once & No & Node& No\\ 
 \hline
 JT-VAE \cite{jin2018junction} & VAE & dataset of graphs & Substructure-Based& No & Node/Edge & No\\
 \hline
 HierVAE \cite{jin2020hierarchical} & VAE & dataset of graphs & Substructure-Based& Yes & Node/Edge & Yes\\
 \hline
 MHG-VAE \cite{kajino2019molecular} & VAE & dataset of graphs & Substructure-Based& No & Node/Edge & No\\
 \hline
 MoleculeChef \cite{bradshaw2019model} & WAE & dataset of graphs & Substructure-Based& No & Node/Edge & No\\
\hline
 CGVAE\cite{liu2018constrained} & VAE & dataset of graphs & Node-by-Node& No & Node/Edge & No\\
   \hline
 DEFactor \cite{assouel2018defactor} & AE & dataset of graphs & Node-by-Node& No & Node/Edge & Yes\\
  \hline
NeVAE\cite{samanta2019nevae}& VAE & dataset of graphs & Node-by-Node& No & Node/Edge & No \\ 
\hline
GraphVRNN \cite{su2019graph}& VAE & dataset of graphs & Node-by-Node & No & Node & No \\ 
\hline
Lim et al. \cite{lim2020scaffold}& VAE & dataset of graphs & Node-by-Node & No & Node/Edge & Yes \\ 
[1ex] 
 \hline
\end{tabular}
}
\caption{\label{tab:autoencoder} The Main Characteristics of Autoencoder-Based Deep Graph Generators}
\end{center}
\end{table*}
\par
GraphVRNN \cite{su2019graph} proposes a VAE-based extension to GraphRNN \cite{you2018graphrnn} to learn the joint probability distributions of graph structure as well as the underlying node attributes by rewriting the likelihood function in Eq. (\ref{eq:2}) as follows:
\begin{equation}
p(S^\pi, X^\pi) = \prod_{i=1}^{n+1} p(S_i^\pi, X_i^\pi|S_{<i}^\pi, X_{<i}^\pi),
\end{equation}
where $X \in \mathbb{R}^{n\times k}$ is the attribute matrix. Then, the authors adopt an autoregressive variational autoencoder to capture the latent factors over graphs with complicated structural dependencies by optimising the lower bound of the likelihood as follows:
\begin{equation}\label{eq:GraphVRNN}
\begin{split}
\mathcal{L}_{\theta,\phi,\psi}(S^\pi, X^\pi) &=
\sum_i \mathop{\mathbb{E}_{\textbf{z}_i\sim q_\psi (.)}}[\log p_\theta(S_i^\pi, X_i^\pi|S_{<i}^\pi, X_{<i}^\pi, \textbf{z}_{\leq i})]\\ &
 -\beta D_{KL}(q_\psi(\textbf{z}_i|S_{\leq i}^\pi, X_{\leq i}^\pi)||p_\phi(\textbf{z}_i|S_{<i}^\pi, X_{<i}^\pi)),
\end{split}
\end{equation}
where $q_\psi(\textbf{z}_i|S_{\leq i}^\pi, X_{\leq i}^\pi)$ and $p_\phi(\textbf{z}_i|S_{<i}^\pi, X_{<i}^\pi)$ are the proposal and the prior distributions in conditional VAE (CVAE) \cite{sohn2015learning} formulation, respectively, and $D_{KL}$ is the Kullback-Leibler (KL) divergence that is tuned by the $\beta$ hyperparameter. An overview of the GraphVRNN framework is depicted in Figure \ref{GraphVRNN}.
\par
Lim et al. \cite{lim2020scaffold} utilize a combination of VAE and DeepGMG \cite{li2018learning} for generating molecular graphs with desired properties containing an arbitrary input scaffold in their structure. To this purpose, the encoder uses a variant of the interaction network \cite{battaglia2016interaction}, \cite{gilmer2017neural} to obtain a representation vector $h_G$ for the input graph, which will be further used to parametrize a normal distribution to sample a latent vector $\textbf{z}$. The decoder, on the other hand, takes a scaffold $S$ as input and extends it by making sequential decisions of node and edge additions in the same way as DeepGMG\cite{li2018learning}, except that it also incorporates the latent vector $\textbf{z}$ in the graph propagation process so that updating node and edge features during the generation is directly affected by $\textbf{z}$. Moreover, the model makes it possible to conduct the process towards generating molecules with desired properties by concatenating the corresponding condition vector $C$ with $\textbf{z}$. After the training with the VAE objective finishes, one could give a scaffold $S$ as well as a condition vector $C$ concatenated with a $\textbf{z}$ sampled from the standard normal distribution to the decoder in order to get a generated molecule with optimized properties.
\section{RL-Based Deep Graph Generators}\label{sec:rl}
This section provides a detailed review of generative approaches utilizing reinforcement learning algorithms to induce desired properties in the generated graphs. Table \ref{tab:rl} gives a comparison among these methods from multiple aspects. \par
\begin{figure*}[t!]
\centering
\includegraphics[width=0.85\textwidth]{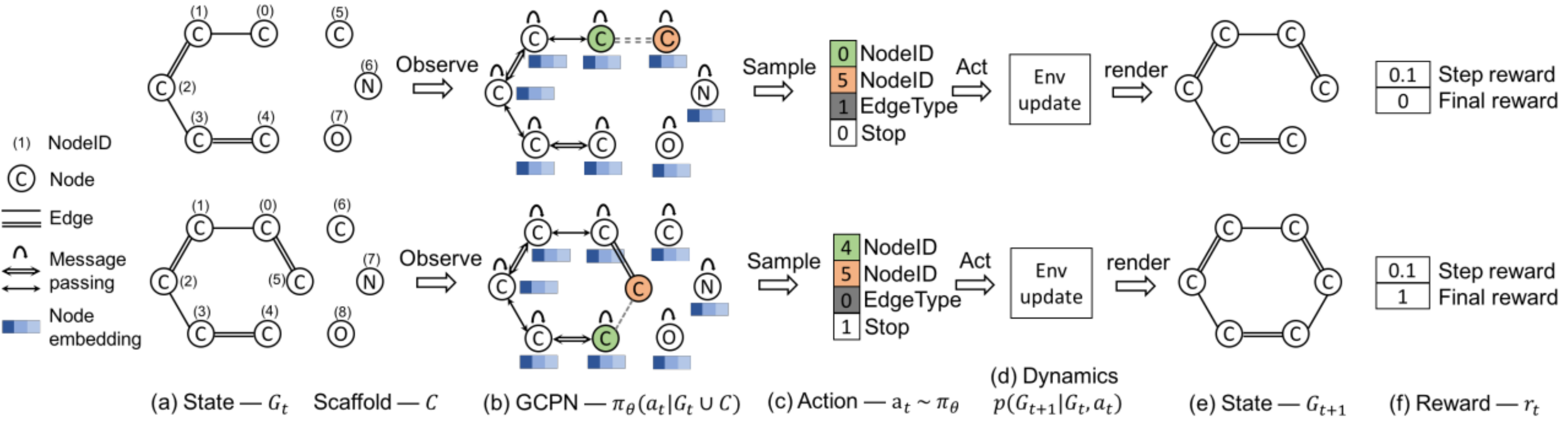}
\vspace*{-0.2cm}
\centering\caption{An overview of GCPN \cite{you2018graph} (reprinted with permission).}
\label{gcpn}
\end{figure*}

GCPN \cite{you2018graph} proposes a stepwise approach for molecular graph generation, formulating the problem as a Markov Decision Process to train an RL agent in a chemistry-aware environment. Thus, at each generation step $t$, the method first takes the intermediate graph $G_t$ and the set of scaffolds $C$ as input and computes the state $s_t$ by applying a GCN variant that supports multiple edge types. It then samples an action $a_t$ from the policy $\pi_\theta$ based on the obtained node embeddings, which can be either to add a new scaffold subgraph or connect two nodes already in the graph. Next, the action will be further processed by the state transition dynamics, and if it violates chemical rules, it will be rejected so that the state stays unchanged. After that, GCPN utilizes two types of rewards to guide the RL agent, namely, intermediate and final rewards, where the former consists of a stepwise validity reward that encourages the generation process to obey chemical valency rules, and an adversarial reward, which employs the GAN framework \cite{goodfellow2014generative} to ensure similarity between real molecules and those to be generated. On the other hand, the final reward includes a domain-specific reward for molecular property optimization and a similar adversarial reward. Finally, the authors adopt Proximal Policy Optimization (PPO) \cite{schulman2017proximal} to optimize the policy network parameters. An overview of the method is depicted in Figure \ref{gcpn}, where each row corresponds to one step in the generation process. 
\par
\begin{figure}[t]
\centering
\includegraphics[width=0.7\textwidth]{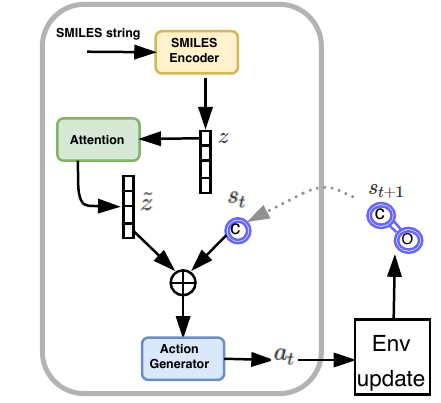}
\vspace*{-1cm}
\begin{center}
\caption{An overview of the framework proposed by Shi et al. \cite{shi2019reinforced}. In the supervised learning phase, only the part in the gray box is trained. On the other hand, the whole architecture is involved in the reinforcement learning stage.}
\label{fig:Shi}
\end{center}
\end{figure}

Subsequently, several methods propose extensions to GCPN. For example, Shi et al. \cite{shi2019reinforced} utilize a combination of general semantic features extracted from the SMILES representations of molecules and their graph representations in order to form more comprehensive states during the generation process. To this end, the authors propose an architecture consisting of a SMILES encoder and an action generator. First, the encoder obtains context vector $z$ from an input SMILES string, which will be further processed by two attention mechanisms, namely action-attention and graph-attention, to get the enhanced context vector $\tilde{z}$. Then, the model concatenates the current graph state $s_t$ with $\tilde{z}$ to pass a heterogeneous state into the action generator that has the same generation mechanism as GCPN, except that it does not involve adversarial rewards. Furthermore, the model is trained in two stages. The supervised learning stage learns an initialization for the model parameters to alleviate the instability of an RL agent training by minimizing the following objective function:
\begin{equation}
J =-\frac{1}{M}\sum_{m=1}^M \log \frac{1}{N} \sum_{n=1}^N\sum_{t}\log P(a_t|\tilde{z}, s_t)+D_{KL}(P_z||P_0),
\end{equation}
where $M$ is the number of molecules in the training dataset, $N$ denodes the number of sampled trajectories for generating each molecule, and $P_z$ and $P_0$ are the distribution of the learned context vector and a prior distribution, respectively. Afterwards, the reinforcement learning stage further optimizes the process towards generating molecular graphs with desired properties. Figure \ref{fig:Shi} provides an overview of this framework.\par
Karimi et al. \cite{karimi2020network} propose another extension to GCPN for drug-combination design, which is a key part of combination therapy. To this end, the authors first develop Hierarchical Variational Graph Auto-Encoders (HVGAE) to embed prior knowledge such as gene-gene, gene-disease, and disease-disease networks to acquire more accurate disease representations. Then, they formulate the problem as generating a set of graphs $\mathcal{G}=\{G^{(k)}\}_{k=1}^K$  conditioned on the learned disease representations by employing a similar generation strategy as GCPN, with the difference that in addition to chemical validity and adversarial rewards, they design a reward to encourage generating disease-specific drug combinations.\par
More recently, DeepGraphMolGen \cite{khemchandani2020deepgraphmolgen} combines GCPN with a molecular property prediction network implemented as a GCN followed by a feedforward layer, which provides GCPN with an extra chemical reward to tilt the process towards generating molecules with additional property, i.e., binding potency to dopamine transporters.
\begin{table*}[ht!]
\begin{center}
\small
 \begin{tabular}{||m{0.15\textwidth} l l l l m{0.12\textwidth} m{0.23\textwidth}||}
 \hline
 Method & \makecell[l]{Generation \\ Strategy} & \makecell[l]{Attention\\ Mechanism}& Features & \makecell[l]{Conditional\\ Generation} & Action & Reward\\ [0.5ex] 
 \hline\hline
 GCPN \cite{you2018graph} & Sequential & No & Node/Edge & No & Link prediction & Domain-specific + GAN\\ 
 \hline
 Shi et al.\cite{shi2019reinforced} & Sequential & Yes & Node/Edge & No & Link prediction & Domain-specific\\ 
 \hline
 Karimi et al. \cite{karimi2020network} & Sequential & Yes & Node/Edge & Yes & Link prediction & Domain-specific + GAN + A reward for drug combinations\\ 
 \hline
 DeepGraphMolGen\cite{khemchandani2020deepgraphmolgen} & Sequential & No & Node/Edge & No & Link prediction & Domain-specific + GAN + Property reward (by the property prediction network)\\ 
 \hline
 GraphOpt \cite{trivedi2020graphopt} & Sequential & No & Node/Edge & No & Link prediction & Learning a reward function via inverse reinforcement learning\\ 
 \hline
 MNCE-RL \cite{xu2020reinforced} & Sequential & No & Node/Edge & No & Production rule selection & Domain-specific + A reward regarding the number of generation steps\\ 
 \hline
 GEGL \cite{ahn2020guiding} & Sequential & No & Node/Edge & No & Generating a molecule & Domain-specific\\ 
[1ex] 
 \hline
\end{tabular}
\caption{\label{tab:rl} The Main Characteristics of RL-Based Deep Graph Generators}
\end{center}
\end{table*}
\par
Besides GCPN and its subsequent approaches, there exist other RL-based methods that generate graph structures by taking different strategies. GraphOpt \cite{trivedi2020graphopt} models graph formation via a Markov Decision Process, aiming to learn both a graph construction procedure $\Pi$ and a  usually unknown latent objective function $\mathcal{F}:G\rightarrow \mathbb{R}$ that reflects the underlying graph formation mechanism. Therefore, inspired by \cite{ng2000algorithms}, the authors formulate the following objective:
\begin{equation}
\begin{split}
\Pi^* &= \mathop{\text{argmin}}_{\Pi} \mathop{\max}_{\mathcal{F}}[\mathcal{F}(G)-\mathcal{F}(\Pi(V))],\\
\mathcal{F}_{opt} &= \mathop{\text{argmax}}_{\mathcal{F}} [\mathcal{F}(G)-\mathcal{F}(\Pi^*(V))],
\end{split}
\end{equation}
where $\mathcal{F}_{opt}$ assigns the highest score to the observed graphs compared to all other ones, and optimization over $\mathcal{F}$ is, in fact, a search for the reward function via inverse reinforcement learning (IRL) \cite{finn2016guided}. In other words, GraphOpt learns a reward function, which is in contrast to most of the RL frameworks that utilize an existing one. The optimal construction procedure, on the other hand, tries to construct a graph $G' = \Pi^*(V)$ in a sequential link formation process given node-set $V$, which is expected to be the most similar graph to the observed one using $\mathcal{F}$ as the similarity measure. To this end, the authors propose a continuous latent action space to infer a link formation action $a_t$ at each time step $t$ by first sampling two vectors $a^{(1)}$ and $a^{(2)}$ from a normal distribution parametrized based on the current graph state $s_t$, which is computed by a GNN \cite{scarselli2008graph}. They then choose two graph nodes with the most similar embeddings to the obtained vectors to construct an edge.\par 
MNCE-RL \cite{xu2020reinforced} proposes a graph convolutional policy network with a novel GCN architecture for generating molecules with optimized properties, which, similar to MHG-VAE\cite{kajino2019molecular}, utilizes grammar rules to guarantee the validity of molecules. To this end, the authors first extend the NCE graph grammar \cite{janssens1982graph} to make it applicable for generating molecules. They then infer the production rules of the grammar from a set of input molecules. Next, the RL-based generation process starts whose action space consists of the set of legal production rules, and at each step, the policy samples a rule based on the node features obtained by applying the proposed GCN on the intermediate graph. A domain-specific reward guides the process towards generating desirable molecules. Moreover, MNCE-RL assigns a negative reward when the number of steps exceeds a threshold to avoid prolonging the generating process. \par
GEGL \cite{ahn2020guiding} proposes to incline a deep neural network called neural apprentice policy towards generating molecules with desired properties. In this respect, the apprentice policy first generates a set of molecules by a SMILES-based LSTM and stores them into a fixed-size max-reward priority queue $\mathcal{Q}$. Then, a genetic expert policy utilizes the content of $\mathcal{Q}$ as seed molecules and applies two genetic operators, namely, the graph-based mutation and crossover \cite{jensen2019graph}, to them and stores the generated molecules in another priority queue denoted by $\mathcal{Q}_{ex}$. After generating each sample, both $\mathcal{Q}$ and $\mathcal{Q}_{ex}$ are updated so that they always contain molecules with the highest rewards. Next, the apprentice policy updates its model's parameters by learning to imitate the molecules stored in $\mathcal{Q}\cup\mathcal{Q}_{ex}$, and the whole procedure repeats iteratively. This way, the expert policy guides the apprentice policy to generate molecules with preferred properties. 
\section{Adversarial Deep Graph Generators}\label{sec:adversarial}
This section reviews methods employing generative adversarial networks (GANs)\cite{goodfellow2014generative} to generate either molecular or non-molecular graph structures. To conduct a more accurate study, we divide the existing approaches into multiple subsections. Moreover, Table \ref{tab:adversarial} provides a multifaceted comparison of them.
\subsection{Random Walk-Based Methods}
A series of works focus on generating random walks instead of the entire graph, as graph random walks are invariant under node reordering. In this respect, NetGAN \cite{bojchevski2018netgan} introduces the first implicit generative model for graphs that learns the distribution of biased random walks over a single graph using the WGAN framework \cite{arjovsky2017wasserstein}. In particular, NetGAN first samples a collection of random walks using the biased second-order strategy \cite{grover2016node2vec} to prepare the model’s training data. Then, the generator learns to sequentially generate random walks node-by-node using the LSTM \cite{hochreiter1997long} architecture, which is initialized by a latent vector $\textbf{z}$ sampled from a standard normal distribution. Meanwhile, the discriminator decides whether a random walk is real or not after processing its entire node sequence by another LSTM. After training finishes, the authors construct the adjacency matrix of a new graph using multiple generated random walks. Further to this, a number of generative approaches have been proposed inspired by the idea of NetGAN or extending it. For example, STGGAN \cite{zhang2019stggan} adopts a similar generating scheme for spatial-temporal graphs.\par
MMGAN \cite{gamage2020multi} generalizes NetGAN to capture higher-order connectivity patterns by introducing multiple types of random walks, each biased towards different motif structures. To simplify the process, MMGAN focuses only on 3-node motifs and proposes an architecture consisting of three GANs, namely, NetGAN that considers pairwise relationships, and two other motif-based GANs. The random walks generated by each of the three GANs are then combined to construct the output graph.\par
{\large S}HADOW{\large C}AST \cite{tann2020shadowcast} proposes another extension to NetGAN in order to make the generation process controllable, which can be considered as a step towards generating graphs with more explainable properties. To this end, the authors first define a graph called $shadow$ with the same structure as the original one but with different node labels so that these node-level properties can control the generation process. Then, they expand the architecture of NetGAN by adding a sequence-to-sequence model called shadow caster, which is implemented by an LSTM \cite{hochreiter1997long}. Specifically, the shadow caster takes in sampled walks from the shadow network and generates synthetic shadow walks of preferred distribution to control the generation process. Next, these model-generated shadow walks are fed into both generator and discriminator as conditions, which are finally trained using the conditional GAN \cite{mirza2014conditional} framework.
\subsection{Graph-Based Methods}
\begin{table*}[ht!]
\begin{center}
\small
 \begin{tabular}{||l l l l l l||}
 \hline
 Method & Input & \makecell[l]{Generation\\ Strategy}& \makecell[l]{Attention\\ Mechanism}& Features& \makecell[l]{Conditional\\ Generation}\\ [0.5ex] 
 \hline\hline
 NetGAN \cite{bojchevski2018netgan} & one single graph & Sequential & No & - & No\\ 
 \hline
 MMGAN \cite{gamage2020multi} & one single graph & Sequential & No & - & No\\
 \hline
 {\large S}HADOW{\large C}AST \cite{tann2020shadowcast} & one single graph & Sequential & No & Node & Yes\\
 \hline
 MolGAN \cite{de2018molgan} & dataset of graphs & All at Once & No & Node/Edge & No\\ 
 \hline
 CONDGEN \cite{yang2019conditional} & dataset of graphs & All at Once & No & - & Yes\\
 \hline
 TSGG-GAN \cite{yang2020learn} & dataset of graphs & All at Once & No & Node & Yes\\
 \hline
 VJTNN + GAN \cite{jin2018learning} & dataset of graphs & Sequential & Yes & Node/Edge & No\\
 \hline
Mol-CycleGAN \cite{maziarka2020mol} & dataset of graphs & Sequential & No & Node/Edge & No\\
 \hline
 Misc-GAN \cite{zhou2019misc} & one single graph & Not mentioned & No & - & No\\
 \hline
\end{tabular}
\caption{\label{tab:adversarial} The Main Characteristics of Adversarial Deep Graph Generators}
\end{center}
\end{table*}
Unlike random walk-based approaches, most of the existing adversarial graph generators deal with the entire graph. Here, we study these methods in two categories.
\subsubsection{General Graph-Based Adversarial DGGs}
\begin{figure}[t]
\centering
\includegraphics[width=0.75\textwidth]{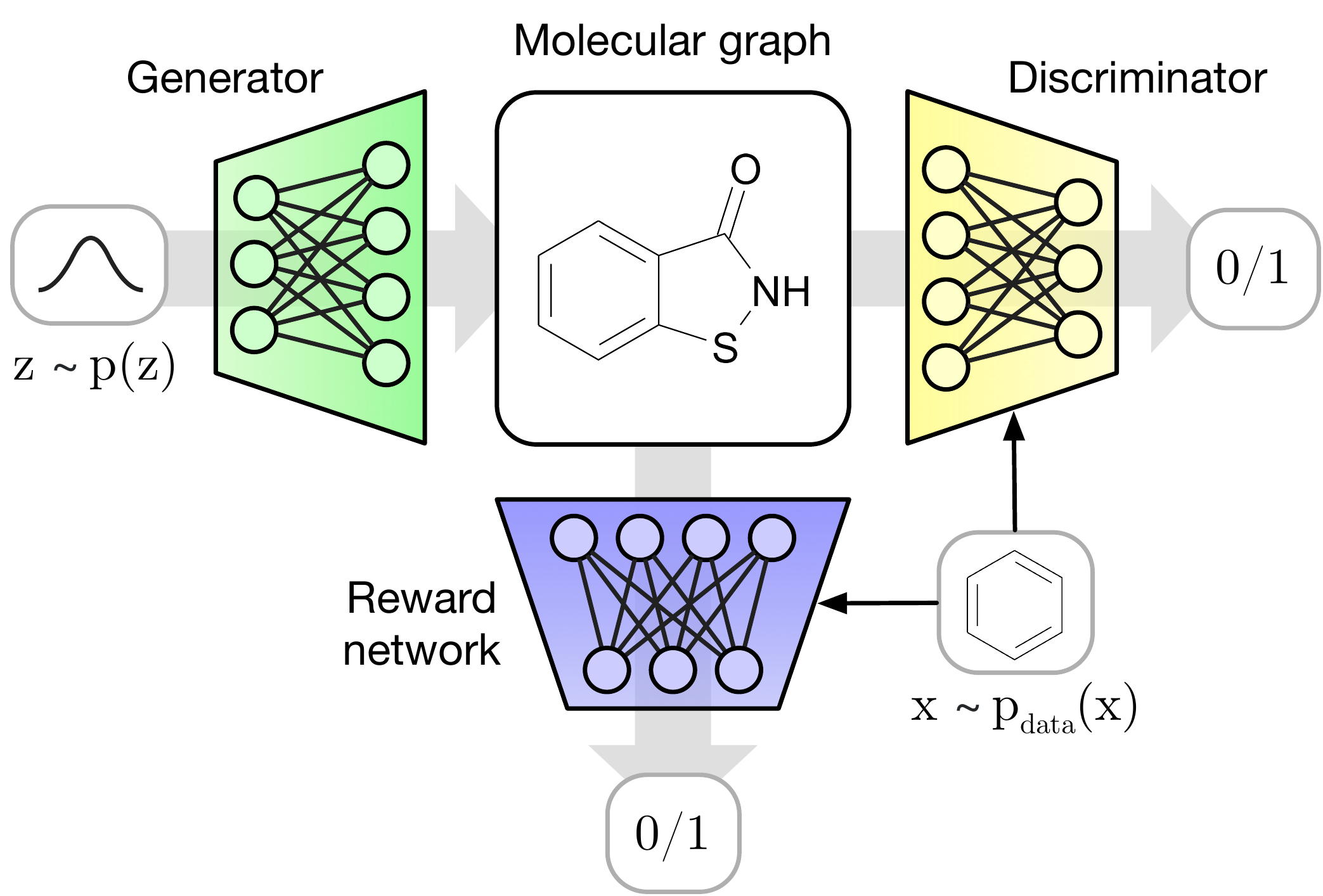}
\vspace*{-1cm}
\begin{center}
\caption{An overview of MolGAN \cite{de2018molgan} (reprinted with permission).}
\label{fig:MolGAN}
\end{center}
\end{figure}

MolGAN \cite{de2018molgan} proposes the first implicit generative model for small molecular graphs. In this respect, its generator first takes a latent vector \textbf{z} sampled from $\mathcal{N}(0, I)$. Then, it outputs a probabilistic graph all at once using an MLP in a way similar to GraphVAE \cite{simonovsky2018graphvae}, which, as a consequence, limits the model to generate graphs of a predefined maximum size. However, in contrast to GraphVAE, MolGAN does not need to perform an expensive graph matching algorithm, as it makes the model likelihood-free using the GAN framework. Next, a permutation-invariant discriminator tries to distinguish between generated graphs and real ones using a combination of the Relational-GCN \cite{schlichtkrull2018modeling} and an MLP. The authors train the discriminator using the WGAN \cite{arjovsky2017wasserstein} objective, while they combine a reinforcement learning objective with that of the WGAN to train the generator, aiming at inclining the process towards generating molecules with desired chemical properties. More precisely, the authors employ a deterministic policy gradient algorithm, namely, DDPG \cite{lillicrap2015continuous}, to maximize the reward, which is approximated by a reward network with the same architecture as the discriminator.  The overall architecture of MolGAN is shown in Figure \ref{fig:MolGAN}.\par 
LGGAN \cite{fan2019conditional} adopts a similar generator to that of MolGAN. However, its discriminator uses JK-Net \cite{xu2018representation} to compute graph embeddings and outputs both the graph label and the probability of the graph being real. Moreover, to incorporate the class information, the authors utilize the AC-GAN \cite{odena2017conditional} framework.
\par
CONDGEN \cite{yang2019conditional} proposes a model of graph variational generative adversarial nets for conditional structure generation. It addresses both the challenges of permutation-invariance and context-structure conditioning indicated by the information of attributes or labels in the networks. The method first applies the trick of latent space conjugation to the base VGAE \cite{kipf2016variational} model in order to convert its node-level encoding into a permutation-invariant graph-level one that allows learning from a dataset of graphs with variable sizes, which is a notable improvement over the VGAE. More precisely, $\mu$ and $\sigma$ in Eq. (\ref{VGAE-1}) are replaced by the following parameters:
\begin{equation}
q(\textbf{z}_{i}|X,A)=\mathcal{N}(\bar{\textbf{z}}|\bar{\mu},\text{diag}(\bar{\sigma}^2)),
\end{equation}
where $\bar{\mu} =\frac{1}{n} \sum_{i=1}^n g_\mu (X,A)_i$ and $\bar{\sigma}^2 =\frac{1}{n^2} \sum_{i=1}^n g_\sigma (X,A)_i^2$. However, the process is still not completely permutation-invariant because the reconstruction loss of the VGAE is computed between the generated adjacency matrix $A'$ and the original matrix $A$, which may be under different node permutations. Therefore, the authors propose a GCN-based discriminator to enforce the structural similarity between the generated and the true adjacency matrices and learn its parameters by optimizing the GAN objective. Thus, the encodings $\text{GCN}_D(A)$ and $\text{GCN}_D(A')$ computed by the discriminator, become permutation-invariant and the reconstruction loss can be computed as $||\text{GCN}_D(A) - \text{GCN}_D(A')||^2_2$. Moreover, according to \cite{mirza2014conditional}, the authors use the concatenation of condition vector $ C $ and latent variable \textbf{Z} to enable conditional structure generation. Then, motivated by CycleGAN \cite{zhu2017unpaired}, they further enforce mapping consistency between the graph context and the structure spaces by sharing the parameters in the two GCN networks, namely, the GCNs in the graph encoder and the discriminator.
\par
More recently, TSGG-GAN \cite{yang2020learn} adopts a time series conditioned generative model that aims to generate a graph given an input multivariate time series, where each time series acts as context information associated with one of the graph nodes. This is particularly the case when it is straightforward to obtain node-level information, while the underlying network is totally unknown. To this end, using SRU \cite{lei2018simple} to extract the information of the time series followed by an MLP, the generator generates the entire graph all at once. At the same time, the discriminator takes a pair of a multivariate time series and a graph as inputs, which are then processed using SRU and GCN, respectively. Thereafter, the discriminator utilizes Neural Tensor Networks (NTN) \cite{socher2013reasoning} to measure the similarity between the time series and the graph and decides whether the graph is real or not.
\subsubsection{Graph-to-Graph Translators}
In addition to the aforementioned methods, there exist other approaches trying to generate a new graph based on an initial one. VJTNN \cite{jin2018learning} proposes a graph-to-graph translation model that learns a mapping from a source molecular graph $X$ to a target graph $Y$ with enhanced chemical properties by utilizing a similar encoder-decoder architecture as JT-VAE \cite{jin2018junction} whose tree decoding process is further enriched by adding an attention mechanism. More specifically, VJTNN augments the basic encoder-decoder model with latent code \textbf{z} derived based on the embeddings of both source and target graphs and minimizes the conditional VAE loss function to learn the mapping $F : (X, \textbf{z}) \rightarrow Y$. The authors then propose an adversarial variation called VJTNN  + GAN to force generated graphs to follow the distribution of the target ones, which is trained using the WGAN framework \cite{arjovsky2017wasserstein}.\par
Mol-CycleGAN \cite{maziarka2020mol} establishes structural similarity between the source and target molecular graphs by adopting a CycleGAN-based \cite{zhu2017unpaired} approach. To this end, the method first computes the latent space embeddings for $X$ and $Y$ using JT-VAE \cite{jin2018junction} and then learns the transformation $F : X \rightarrow Y$ ( and its reverse, i.e., $G : Y \rightarrow X$) in that space. Mol-CycleGAN also introduces the discriminator $D_X$ (and $D_Y$) to decide whether a sample is from the distribution of $X$ (or $Y$) or it is generated by $G$ (or $F$). The model parameters are trained by optimizing the following loss function:
\begin{equation}
\begin{split}
\mathcal{L}(F, G, D_X, D_Y)=&\mathcal{L}_{GAN}(F, D_Y, X, Y)+\mathcal{L}_{GAN}(G, D_X, Y, X)\\
&+\lambda_1\mathcal{L}_{cyc}(F, G) + \lambda_2\mathcal{L}_{identity}(F, G),
\end{split}
\end{equation}
where the authors utilize the adversarial loss of LS-GAN \cite{mao2017least} and the similar $\mathcal{L}_{cyc}(F, G)$ and $\mathcal{L}_{identity}(F, G)$ as CycleGAN, where the former reduces the space of mapping functions and acts as a regularizer, while the latter makes the generated molecule not to be structurally far away from the original one. After the training finishes, Mol-CycleGAN takes a molecule $X$ as input and calculates its embedding by applying the encoder of the JT-VAE. Then, $F(X)$ computes an embedding corresponding to a molecule with desired properties that is also structurally similar to $X$. Finally, the model generates the optimized molecular graph $Y$ using the JT-VAE's decoder.\par
Misc-GAN \cite{zhou2019misc} proposes another translation model inspired by CycleGAN \cite{zhu2017unpaired} to learn a mapping function $F$ from a source graph $G_s$ to its corresponding target graph $G_t$ while preserving the hierarchical graph structures (i.e., the community structures) in the target graph in different levels of granularity. The model training consists of three stages: First, it constructs coarser graphs in $ L $ granularity levels based on an input target graph $G_t$. Then, at each level $l$, it trains an independent CycleGAN-based generative model from $G_s$ to $G_t^{(l)}$. Finally, all generated graphs $\tilde{G}_t^{(l)}$ are aggregated together to form the reconstructed target graph $\tilde{G}_t$. The framework is trained by minimizing the following loss function:
\begin{equation}
\mathcal{L} = \mathcal{L}_{ms} + \mathcal{L}_F + \mathcal{L}_G + \mathcal{L}_{cyc},
\end{equation}
where $\mathcal{L}_{ms}$ is the multi-scale reconstruction loss between the target graph $G_t$ and the generated graph $\tilde{G}_t$, $\mathcal{L}_F$ is the forward adversarial loss for learning a mapping from the source to the target graph, $\mathcal{L}_G$ is the backward adversarial loss to learn the reverse mapping, and $\mathcal{L}_{cyc}$ is the cycle consistency loss \cite{zhu2017unpaired}.
\section{Flow-based Deep Graph Generators}\label{sec:flow}
\begin{figure}[t]
\centering
\includegraphics[width=0.55\textwidth]{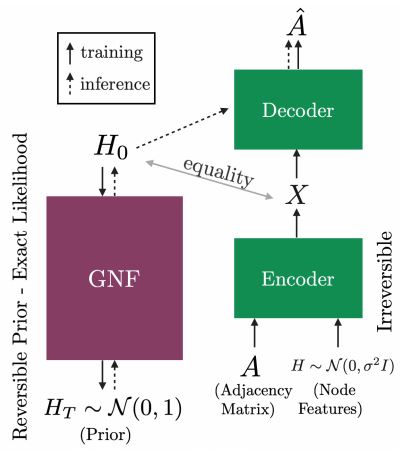}
\vspace*{-1cm}
\begin{center}
\caption{The framework of GNF \cite{liu2019graph} for the graph generation (reprinted with permission).}
\label{fig:GNF}
\end{center}
\end{figure}

\begin{table*}[ht!]
\begin{center}
\small
 \begin{tabular}{||l l l l l l||}
 \hline
 Method & Category & \makecell[l]{Generation\\ Strategy} & \makecell[l]{Attention\\ Mechanism} & Features & \makecell[l]{Conditional\\ Generation}\\ [0.5ex] 
 \hline\hline
 GNF \cite{liu2019graph} & Autoencoder-based & All at Once & Yes& Node/Edge& No\\ 
\hline
GraphNVP \cite{madhawa2019graphnvp} & - & All at Once & No& Node/Edge& No\\ 
\hline
GraphAF \cite{shi2020graphaf} & Autoregressive & Node-by-node & No& Node/Edge& No\\ 
\hline
GrAD \cite{shah2020auto} & Autoregressive & Block of nodes & Yes& -& No\\ 
[1ex] 
 \hline
\end{tabular}
\caption{\label{tab:flow-based} The Main Characteristics of Flow-Based Deep Graph Generators}
\end{center}
\end{table*}
In addition to the methods we have discussed so far, a line of research has recently emerged, which employs flow-based approaches in the field of graph generation. For example, GNF \cite{liu2019graph} develops a generative model of graphs by combining normalizing flows with a graph auto-encoder. More specifically, the authors first train a permutation invariant graph auto-encoder that encodes an input graph to a set of node features $X\in \mathbb{R}^{n\times k}$ using a standard GNN. Then, a simple decoder outputs a probabilistic adjacency matrix $\hat{A}$, in which edge probability between two arbitrary nodes $i$ and $j$ with embedding vectors $x_i$ and $x_j$ is computed as follows:
\begin{equation}
\hat{A}_{ij}=\frac{1}{1+\exp(C(||x_i-x_j||^2_2-1))},
\end{equation}
where C is a temperature hyperparameter. After the auto-encoder training completes, the encoder is employed to compute node features X to be used as training input for the GNF. Then, the GNF, which is based on non-volume preserving flows \cite{dinh2016density}, learns a mapping from the complicated graph distribution into a latent distribution that is well modelled as a Gaussian. At inference time, GNF generates node features by first sampling $Z \sim \mathcal{N} (0, I)$ from the latent space followed by applying the inverse mapping, $X = f^{-1}(Z)$ which is then fed into the decoder to get the predicted adjacency matrix as illustrated in Figure \ref{fig:GNF}. GraphNVP \cite{madhawa2019graphnvp} takes a similar approach, but rather than pretraining an auto-encoder to get continuous node features, the authors propose to perform Dequantization \cite{dinh2016density}, \cite{kingma2018glow} by adding uniform noise to the discrete adjacency tensor as well as the node label matrix. More precisely, GraphNVP proposes a two-step generation scheme by learning two latent representations for each graph, one for the adjacency tensor and the other for node labels. 
\par
In addition to the flow-based methods we have studied so far that generate the whole graph in one step, there are other approaches adopting autoregressive generation strategies. For example, GraphAF \cite{shi2020graphaf} proposes to generate molecular graphs by combining the advantages of both autoregressive and flow-based models. The method first converts a molecular graph structure $G = (A, C)$, where both node type matrix $C$ and the adjacency matrix $A$ are discrete, into continuous data $G' = (A', C')$ using Dequantization technique \cite{dinh2016density}, \cite{kingma2018glow} in order to make the data usable for a flow-based model. Then, conditional distributions for the $i$-th generation step are defined according to Autoregressive Flows (AF) \cite{papamakarios2017masked} as follows:
\begin{equation}\label{GraphA-1}
\begin{split}
&p(C'_i|G_i) =\mathcal{N}(\mu^C_i, (\alpha_i^C)^2)\\
&p(A'_{ij}|G_i, C_i, A_{i, 1:j-1}) =\mathcal{N}(\mu^A_{ij}, (\alpha_{ij}^A)^2),
\end{split}
\end{equation}
where $G_i$ is the current sub-graph, $\mu^C_i$, $\alpha_i^C$, and $\mu^A_{ij}$, $\alpha_{ij}^A$ are the means and standard deviations of Gaussian distributions, which are computed by different neural networks based on node embeddings of the sub-graph generated so far. To calculate the exact likelihood, an invertible mapping from the molecule structures $G' = (A', C')$ to latent Gaussian space \textbf{z} is defined as:
\begin{equation}
z_i=(C'_i-\mu^C_i)\odot\frac{1}{\alpha_i^C}  ,\ z_{ij}= (A'_{ij}-\mu^A_{ij})\odot\frac{1}{\alpha_{ij}^A},
\end{equation}
where $\frac{1}{\alpha_i^C}$ and $\frac{1}{\alpha_{ij}^A}$ denote element-wise reciprocals of $\alpha_i^C$ and $\alpha_{ij}^A$, respectively and $\odot$ is the element-wise multiplication. At inference time, GraphAF just samples random variables $z_i$ and $z_{ij}$ from the latent Gaussian space and converts them to the molecule structures as in Eq. (\ref{GraphA-2}) to generate new graphs in an autoregressive manner:
\begin{equation}\label{GraphA-2}
C'_i=z_i\odot \alpha_i^C+\mu^C_i\ ,\ A'_{ij}=z_{ij}\odot \alpha_{ij}^A+\mu^A_{ij}.
\end{equation}
GraphAF further proposes a valency-based rejection sampling similar to MolecularRNN \cite{popova2019molecularrnn} to guarantee the validity of generated molecules. The authors also propose to fine-tune the generation process with reinforcement learning to generate molecules with optimized properties.
\par
More recently, GrAD \cite{shah2020auto} proposes another autoregressive flow-based approach for graph generation, which can also be considered as a variant of GRAN \cite{liao2019efficient}. In particular, its training consists of two stages. Firstly, for generating each new block of $B$ nodes in the $t$-th step, the model samples latent codes $H_{b_t}\in \mathbb{R}^{B\times k}$ to initialize the corresponding node representations. Then, different from GRAN's formulation in Eq. (\ref{GRAN-GNN}), the node features get updated using graph attention layers almost according to \cite{vaswani2017attention}, except that self-attention weights are calculated only based on each node's neighborhood to inject structural information of the currently generated graph into the updating process. After node representations are obtained, the model follows a similar generation strategy as GRAN and jointly optimizes the generator's parameters and the distribution of latent codes. In the second training stage, GrAD trains a flow-based reversible model to map samples from the optimized distribution of latent codes to a simple Gaussian distribution. Therefore, at the inference time, one can first sample a batch of $B$ vectors $Z\in \mathbb{R}^{B\times k}$ from a Gaussian base distribution and apply the inverse mapping $H_{b_t}=f^{-1}(Z)$ to obtain initial node representations and then go through the generation process.\par
Table \ref{tab:flow-based} summarizes the main characteristics of the flow-based DGGs. 

\section{Applications}\label{sec:applications}
Deep graph generation approaches have a wide range of applications, from discovering new molecular structures and building knowledge graphs to modeling physical, social, and biological networks. Here we review some of the most explored applications and suggest potential future directions.
\subsection{Molecular Graph Generation}
The molecule generation approaches aim to downsize the high-dimensional chemical space to expedite drug design and material discovery. Since molecules can be considered as graphs, where the atoms form the graph's node-set and the chemical bonds determine how those nodes connect, the most widely explored application of modern deep graph generative methods is generating molecular structures, a problem that has been previously mostly formulated as producing SMILES strings\cite{olivecrona2017molecular, kusner2017grammar, dai2018syntax, gomez2018automatic, popova2018deep}. Using graph generators instead of SMILES based models results in generating more valid intermediate substructures compared to mostly meaningless partially generated substrings. It also allows better capturing the similarities between molecules as molecules with similar structures can have totally different SMILES encodings.\par 
Deep molecular graph generation approaches proposed so far utilize various frameworks, from VAEs\cite{flam2020graph, simonovsky2018graphvae, jin2018junction, ma2018constrained, liu2018constrained, samanta2019nevae, bradshaw2019model, kajino2019molecular} and  GANs\cite{de2018molgan, maziarka2020mol, jin2020hierarchical} to RL-based \cite{you2018graph, shi2019reinforced, xu2020reinforced}, autoregressive \cite{shi2020graphaf, kawai2019scalable, li2018multi, popova2019molecularrnn, li2018learning, lim2020scaffold} and flow-based frameworks \cite{liu2019graph, madhawa2019graphnvp, shi2020graphaf}. They also adopt different generation strategies at varying granularity levels. For example, some of them generate the whole graph all at once \cite{simonovsky2018graphvae, de2018molgan}, while others add one atom at a time \cite{li2018multi, popova2019molecularrnn, li2018learning} or use valid chemical substructures as their building blocks \cite{jin2018junction, jin2020hierarchical, lim2020scaffold}.\par 
Moreover, in molecular graph generation, two challenges must be taken into account. First, the generated molecules must satisfy the explicitly specified validity constraints, i.e., an atom's chemical bonds should not exceed its valence. The graph generator models proposed so far address the issue by employing different mechanisms, including introducing structural penalties during the training \cite{popova2019molecularrnn, you2018graph}, adopting valency-based rejection sampling at the inference time \cite{shi2020graphaf, popova2019molecularrnn}, utilizing a grammar-based approach \cite{kajino2019molecular}, adding regularization terms to the objective function\cite{ma2018constrained}, using valid chemical substructures as building blocks\cite{bradshaw2019model, jin2018junction, lim2020scaffold, jin2020hierarchical}, and employing valency masking mechanism\cite{liu2018constrained, samanta2019nevae}. The second challenge to be considered is that new molecular structures should obey some desired properties. This problem has also been addressed by taking various strategies including minimizing a distance \cite{liu2018constrained, bradshaw2019model} or utilizing Bayesian optimization \cite{kajino2019molecular, jin2018junction, samanta2019nevae} in some continuous latent space, maximizing a domain-specific reward in RL-based approaches \cite{you2018graph, shi2019reinforced, de2018molgan}, or performing the generation given an input molecule with desired properties and try to preserve those properties in the target molecule \cite{maziarka2020mol, jin2020hierarchical}.

\subsection{Non-Molecular Graph Generation}
Although the most remarkable application of modern graph generation approaches explored so far is generating molecular structures, several other non-application-specific approaches have been proposed \cite{you2018graphrnn, su2019graph, bacciu2020edge, liao2019efficient, guo2020interpretable, shah2020auto, yang2019conditional} working on more general datasets. While these methods' ultimate goal is to be used on real-world applications such as generating social network graphs, most of them suffer from scalability issues. Thus, they are currently being applied to synthetic or relatively small real datasets. Despite the steps taken towards making these models more scalable \cite{you2018graphrnn, liao2019efficient, dai2020scalable}, it should be specifically considered as future work.
\subsection{Future Applications}
Beyond the discussed applications, some other problems can potentially be solved from a graph generation perspective. This is particularly the case when the output space includes graphs, and so the generated output, on the one hand, must depend on the input and, on the other hand, must obey the distribution of the output space graphs. Below, two practical examples of these problems are mentioned.
\subsubsection{Language-Based Graph Generation}
In natural language processing, several approaches have been proposed to extract rich graph-structured information from textual data. They include methods aiming to extract AMRs (Abstract Meaning Representations) \cite{wang2015transition, lyu2018amr, zhang2019amr}, semantic graphs\cite{chen2018sequence}, semantic dependency graphs\cite{DozatM17, wang2018neural, dozat2018simpler}, and even those that transform one graph into another based on some input sentences\cite{Johnson17}. However, most existing methods often propose some domain-specific procedures to produce these graph-structured knowledge representations, which, as a result, are not capable enough to consider various effective factors. Therefore, as a future orientation, the community can solve such problems with a conditional generative approach, making it possible to generate graphs from their corresponding distribution given the specified textual input.
\subsubsection{Scene Graph Generation}
Scene graphs are structured representations of images providing higher-level knowledge for scene understanding, where the objects in each image form the node-set of the corresponding scene graph, and the relationships between objects determine how the nodes connect. As this class of graphs has a wide range of applications, including image captioning, visual question answering, image retrieval, and image generation \cite{gu2019scene}, scene graph generation becomes a line of research in recent years. \par
Most scene graph generation methods first detect objects from an input image using object detection models like Faster R-CNN \cite{ren2015faster} to form the set of graph nodes. Meanwhile, the relationships can be extracted either jointly with the objects \cite{li2017scene} or after all the objects are detected \cite{xu2017scene, yang2018graph, qi2019attentive, chen2019knowledge}. Among these approaches, some generate scene graphs solely based on input images \cite{newell2017pixels, xu2017scene, yang2018graph}, while others benefit from additional text input \cite{qi2019attentive, khademi2020deep}, or some self-generated extra information \cite{gu2019scene, chen2019knowledge}. Moreover, as the number of objects increases, some models \cite{li2018factorizable, klawonn2018generating, gu2019scene} propose to first generate multiple subgraphs and then aggregate them to construct the complete scene graph to address the scalability issue.\par 
Here, we have briefly reviewed and categorized some of the scene graph generation methods. However, they are more of a relational information extractor from images rather than graph generators, which estimate the underlying data distribution. Therefore, it can be explored in the future.
\section{Implementations}\label{sec:implementations}
In this section, we discuss the implementation details by categorizing and summarizing commonly used datasets and evaluation metrics. We also collect the available source codes in Appendix A.
\subsection{Datasets}
There are many datasets for learning on graphs that have been investigated by previous studies, including \cite{wu2020comprehensive} and \cite{hu2020open}. However, none of these studies has thoroughly and exclusively examined and categorized the datasets used in graph generation approaches. Here, we have summarized the most prominent ones in three general categories according to the main graph generation applications, as shown in Table \ref{tab:Dataset}.\\
\begin{table}[ht!]
\small
\begin{tabular}{||l|m{0.19\textwidth}|m{0.4\textwidth}||}
\hline
Category                                                                                 & Dataset         & Citation \\ \hline \hline
\multirow{18}{*}{\begin{tabular}[c]{@{}l@{}}Chemical\\ \&\\ Bioinformatics\end{tabular}} & ZINC            &   \cite{xu2020reinforced, popova2019molecularrnn, goyal2020graphgen, kawai2019scalable, shi2020graphaf, liu2018constrained, you2018graph, shi2019reinforced, assouel2018defactor,khemchandani2020deepgraphmolgen, ahn2020guiding, jin2018junction,ma2018constrained,simonovsky2018graphvae, kajino2019molecular,samanta2019nevae, jin2018learning, maziarka2020mol, madhawa2019graphnvp}
       \\ \cline{2-3} 
                                                                                         & QM9             &   \cite{shi2020graphaf, simonovsky2018graphvae, flam2020graph, ma2018constrained, liu2018constrained, samanta2019nevae,liu2019graph, de2018molgan, madhawa2019graphnvp}       \\ \cline{2-3} 
                                                                                         & CEPDB           & \cite{liu2018constrained}         \\ \cline{2-3} 
                                                                                         & Polymer         &   \cite{jin2020hierarchical}       \\ \cline{2-3} 
                                                                                         & USPTO           &  \cite{bradshaw2019model}        \\ \cline{2-3} 
                                                                                         & GuacaMol        &   \cite{ahn2020guiding, xu2020reinforced}     \\ \cline{2-3} 
                                                                                         & ChEMBL          &  \cite{li2018multi, popova2019molecularrnn, li2018learning, maziarka2020mol}        \\ \cline{2-3} 
                                                                                         & Protein         &   \cite{you2018graphrnn, bacciu2020edge, liao2019efficient, kawai2019scalable, dai2020scalable,shah2020auto, guo2020interpretable}      \\ \cline{2-3} 
                                                                                         & MOSES           &    \cite{popova2019molecularrnn, shi2020graphaf}      \\ \cline{2-3} 
                                                                                         & NCI-H23         &  \cite{goyal2020graphgen}        \\ \cline{2-3} 
                                                                                         & Yeast           &     \cite{goyal2020graphgen}     \\ \cline{2-3} 
                                                                                         & MOLT-4          &     \cite{goyal2020graphgen}    \\ \cline{2-3} 
                                                                                         & MCF-7           &     \cite{goyal2020graphgen}     \\ \cline{2-3} 
                                                                                         & Enzymes         &    \cite{bacciu2020edge, goyal2020graphgen}     \\ \cline{2-3} 
                                                                                         & PPI             &   \cite{liu2019graph}       \\ \hline
\multirow{5}{*}{Social}                                                                  & Cora            &    \cite{goyal2020graphgen, kipf2016variational, grover2019graphite, liu2019graph, bojchevski2018netgan,gamage2020multi, tann2020shadowcast, trivedi2020graphopt}      \\ \cline{2-3} 
                                                                                         & Citeseer        &   \cite{goyal2020graphgen, kipf2016variational, grover2019graphite, bojchevski2018netgan, gamage2020multi, trivedi2020graphopt}       \\ \cline{2-3} 
                                                                                         & Pubmed          &   \cite{kipf2016variational, grover2019graphite, liu2019graph, bojchevski2018netgan, trivedi2020graphopt}     \\ \cline{2-3} 
                                                                                         & DBLP            &    \cite{bojchevski2018netgan, yang2019conditional}      \\ \hline
\multirow{11}{*}{Synthetic}                                                               & Barabasi-Albert &     \cite{you2018graphrnn, li2018learning, kawai2019scalable, grover2019graphite, yang2020learn, trivedi2020graphopt, li2020dirichlet}  \\ \cline{2-3} 
                                                                                         & Erdos-Renyi     &   \cite{grover2019graphite, guo2020interpretable, trivedi2020graphopt, li2020dirichlet}       \\ \cline{2-3} 
                                                                                         & Watts-Strogatz  &  \cite{guo2020interpretable}        \\ \cline{2-3} 
                                                                                         & Community       &  \cite{you2018graphrnn, su2019graph, bacciu2020edge, kawai2019scalable, shi2020graphaf,shah2020auto, liu2019graph}        \\ \cline{2-3} 
                                                                                         & Grid            &    \cite{you2018graphrnn, liao2019efficient, kawai2019scalable, dai2020scalable, shah2020auto}     \\ \cline{2-3} 
                                                                                         & Lobster         &  \cite{kawai2019scalable, dai2020scalable, shah2020auto}       \\ \cline{2-3} 
                                                                                         & Cycles          &    \cite{li2018learning, shah2020auto}      \\ \cline{2-3} 
                                                                                         & Ego             &  \cite{you2018graphrnn, su2019graph, bacciu2020edge, kawai2019scalable, fan2020attention, shi2020graphaf, shah2020auto, grover2019graphite, liu2019graph, li2020dirichlet}        \\ \hline
\end{tabular}
\caption{\label{tab:Dataset} Summary of the Commonly Used Datasets}
\end{table}
\subsection{Evaluation Metrics}
Depending on the application, graph generation approaches use different evaluation metrics. Specifically, the molecular graph generators adopt two different sets of metrics where the first set contains those evaluating the overall quality of generated samples, including \textit{validity}, \textit{uniqueness}, \textit{novelty}, \textit{reconstruction}, \textit{internal diversity}, \textit{negative log-likelihood (NLL)}, and some structural statistics like \textit{nearest neighbor similarity (SNN)} or \textit{fragment/scaffold similarity}. The second set, on the other hand, contains metrics assessing special chemical properties of the molecules, namely, \textit{synthetic accessibility score (SA score)}\cite{ertl2009estimation}, \textit{drug-likeness score (QED)}\cite{bickerton2012quantifying}, \textit{molecular weight (MW)}, \textit{log partition coefficient (logP)}, \textit{penalized logP}, and \textit{topological polar surface area (TPSA)}. For non-molecular graph generation, on the other side, a considerable number of approaches employ distribution-related metrics such as \textit{Kullback-Leibler Divergence (KLD)} or \textit{Maximum Mean Discrepancy (MMD)} on several graph statistics like degrees, clustering coefficients, orbit counts, and the spectra of the graphs from the eigenvalues of the normalized graph Laplacian. \textit{NLL}, \textit{validity}, \textit{novelty}, and \textit{uniqueness} are also among other metrics adopted to evaluate non-molecular approaches.
\section{Future Directions}\label{sec:future}
Although several deep graph generation models have been proposed in the past few years, due to the emergence of this field and its short history, a number of challenges remain, suggesting future directions for research, as follows.
\subsection{Scalability}
Most of the proposed graph generation methods are only applicable to small graphs with a maximum of a few tens of nodes. Therefore, designing molecular graphs, which are mostly small in size, is the most prominent application of DGGs so far. Although some initial steps \cite{you2018graphrnn, liao2019efficient, dai2020scalable} have been taken towards scalability of generator models, much more effort is necessary to make them applicable in a wider range of real-world applications, such as social network modeling.
\subsection{node ordering}
Each graph with $n$ nodes can be represented under $n!$ different node orderings. Therefore, it becomes intractable for likelihood-based DGGs to calculate the exact likelihood as the graph size increases. To address this issue, some approximate approaches such as using approximate graph matching algorithms \cite{simonovsky2018graphvae} or maximizing a lower bound of the likelihood by only considering subsets of node orderings (i.e., fixed \cite{li2018learning, dai2020scalable}, uniform random \cite{li2018learning}, BFS \cite{you2018graphrnn}, or a family of canonical orderings \cite{liao2019efficient}), have been proposed. However, it is still necessary to provide more effective solutions for the node ordering problem, as a result of which, the generators can generate larger samples with higher quality.
\subsection{Interpretability}
As mentioned earlier, DGGs are utilized in critical applications such as designing drug molecules, which directly affects public health. Therefore, the more transparent the generation procedure, the better control is exercised on the desirability of generated samples, which prevents additional trials and errors by limiting the number of candidate solutions. Hence, it is of great importance to make graph generation methods more interpretable. As of now, deep generative methods in areas such as image\cite{biffi2018learning, voynov2019rpgan, biffi2020explainable} and text \cite{wen2017latent, zhao2018unsupervised,  shi2020dispersed} have slowly moved towards being more interpretable. However, only a few attempts \cite{stoehr2019disentangling, guo2020interpretable, li2020dirichlet} have recently been made in graph generation, making model interpretability a notable future research prospect.
\subsection{Dynamic Graphs}
While existing DGGs focus on generating static graphs, most of the graphs are inherently dynamic, meaning that they change over time by earning/losing nodes or connections, or even their attributes may alter. For example, in a social network, some users may join/leave the network, or the relationships between existing users may change over time. Therefore, generating dynamic graphs would play a key role in predicting how networks evolve. However, dynamicity is almost not addressed in the current generative approaches, making it a potentially challenging problem to explore in the further. 
\subsection{Conditional Graph Generation}
    When generating new graphs, in most cases, one aims to discover structures with desired characteristics. While conditional generation is relatively well investigated in image \cite{odena2017conditional, van2016conditional, yan2016attribute2image, wang2018high} and text \cite{hu2017toward, zhou2018emotional, keskar2019ctrl} domains, it is comparably less explored in the field of graph generation. For example, in molecular graph generation, the generated molecules must satisfy some validity constraints or hold desired chemical properties. However, only a limited number of proposed methods adopt a conditional approach by whether incorporating conditional codes into the generation process \cite{lim2020scaffold, simonovsky2018graphvae}, enforcing the existence of favourable substructures in the output graph \cite{lim2020scaffold} or performing the generation conditioned on an input molecule to ensure the structural similarity \cite{jin2020hierarchical}. Meanwhile, the majority of methods do not formulate the problem as conditional generation and address the issue by employing other techniques like property optimization in some latent continuous space \cite{liu2018constrained, bradshaw2019model, kajino2019molecular, jin2018junction, samanta2019nevae} or injecting validity constraints, whether at the training \cite{popova2019molecularrnn, ma2018constrained} or the inference time \cite{shi2020graphaf, popova2019molecularrnn}. Nevertheless, this issue has been even less studied in the non-molecular graph generation models, and only a few of them have partially addressed the problem \cite{yang2019conditional, tann2020shadowcast}. Therefore, focusing more on conditional graph generation problems, especially those that have not yet been explored, such as class conditioned generation, is an important future research direction.
\section{Conclusion}\label{sec:conclusion}
In this article, we surveyed the emerging field of deep learning-based graph generation. For this purpose, we classified the existing methods into five general categories. We then provided a detailed and comparative review of the approaches in each category. We summarized the implementation details, including datasets, evaluation metrics, and available source codes, and discussed the current applications and possible future trends. Finally, we suggested future research directions according to the current challenges. We believe this article provides the readers a comprehensive insight to the field of graph generation research.

\appendices
\section{Acronym}
Table \ref{tab:acronyms} summarizes the acronyms and nomenclature used in this survey.
\begin{table}[ht!]
\small
\begin{tabular}{||m{0.18\textwidth}|m{0.5\textwidth}|l||}
\hline
Acronym      & Model Name                                   & Reference \\ \hline \hline
DGG          & Deep Graph Generator                    &           \\ \hline
RNN          & Recurrent Neural Network                    &           \\ \hline
LSTM          & Long Short-Term Memory                    & \cite{hochreiter1997long}          \\ \hline
GRU          & Gated Recurrent Unit                    & \cite{cho2014learning}          \\ \hline
VAE          & Variational Autoencoder                     &\cite{kingma2013auto}\\ \hline
GAN          & Generative Adversarial Network               &\cite{goodfellow2014generative}\\ \hline
GNN          & Graph Neural Network               &\cite{scarselli2008graph}\\ \hline
GCN          & Graph Convolutional Network               &\cite{kipf2016semi}\\ \hline
REIN        &       Recurrent Edge Inference Network                                      &\cite{daroya2020rein}\\ \hline
DeepNC        &       Deep Generative Network Completion                          &\cite{tran2019deepnc}\\ \hline
GRAN         & Graph Recurrent Attention Network            &\cite{liao2019efficient}\\ \hline
GRAM         & Graph Generative Model with Graph Attention Mechanism                    &\cite{kawai2019scalable}\\ \hline
AGE          & Attention-Based Graph Evolution              &\cite{fan2020attention}\\ \hline
DeepGMG      &          Deep Generative Models of Graphs                                    &\cite{li2018learning}\\ \hline
DeepGG      &          Deep Graph Generators          &\cite{stier2020deep}\\ \hline
BiGG         & BIg Graph Generation                         &\cite{dai2020scalable}\\ \hline
VGAE         & Variational Graph Autoencoder               &\cite{kipf2016variational}\\ \hline
MPGVAE       & Message Passing Graph VAE                    &\cite{flam2020graph} \\ \hline
NED-VAE      & Node-Edge Disentangled VAE                   &\cite{guo2020interpretable}\\ \hline
DGVAE      & Dirichlet Graph VAE                   &\cite{li2020dirichlet}\\ \hline
JT-VAE       & Junction Tree VAE                            &\cite{jin2018junction}\\ \hline
HierVAE      & Hierarchical VAE                             &\cite{jin2020hierarchical}\\ \hline
MHG-VAE      & Molecular Hypergraph Grammar VAE             &\cite{kajino2019molecular}\\ \hline
CGVAE        & Constrained Graph VAE                        &\cite{liu2018constrained}\\ \hline
DEFactor     & Differentiable Edge
Factorization-based Probabilistic Graph
Generation      &\cite{assouel2018defactor}\\ \hline
GraphVRNN    & Graph Variational RNN   &\cite{su2019graph}\\ \hline
GCPN         & Graph Convolutional Policy Network           &\cite{you2018graph}\\ \hline
MNCE-RL         & Molecular Neighborhood-Controlled Embedding RL               &\cite{xu2020reinforced}\\ \hline
GEGL         & Genetic Expert-Guided Learning               &\cite{ahn2020guiding}\\ \hline
MMGAN        & Multi-MotifGAN                               &\cite{gamage2020multi}\\ \hline
MolGAN       &    Molecular GAN &\cite{de2018molgan}\\ \hline
LGGAN       &    Labeled Graph GAN &\cite{fan2019conditional}\\ \hline
TSGG-GAN     & Time Series Conditioned Graph Generation-GAN &\cite{yang2020learn}\\ \hline
VJTNN     & Variational Junction Tree Encoder-Decoder &\cite{jin2018learning}\\ \hline
Misc-GAN     & Multi-Scale GAN                              &\cite{zhou2019misc}\\ \hline
GNF      & Graph Normalizing Flow       &\cite{liu2019graph}\\ \hline
GrAD         & Graph Auto-Decoder                           &\cite{shah2020auto}\\ \hline
\end{tabular}
\caption{\label{tab:acronyms} Acronyms with their extended names}
\end{table}
\section{Source Codes}
Table \ref{tab:codes} summarizes the set of publicly available source codes for deep learning-based graph generation approaches discussed in the survey.
\begin{table*}
\begin{center}
\small
\begin{tabular}[t]{||l|lm{0.33\textwidth}ll||}
\hline
Category                         & Method        & URL & Language/Framework & O.A.\\ \hline \hline
\multirow{13}{*}{Autoregressive} & MolMP\cite{li2018multi}         & \url{https://github.com/kevinid/molecule_generator}  &   Python/MXNet     & Yes            \\ \cline{2-5} 
                                 & MolRNN\cite{li2018multi}        &   \url{https://github.com/kevinid/molecule_generator}  &  Python/MXNet   & Yes               \\ \cline{2-5} 
                                 & GraphRNN\cite{you2018graphrnn}    &   \url{https://github.com/JiaxuanYou/graph-generation}  &  Python/PyTorch & Yes                 \\ \cline{2-5} 
                                                                  & DeepNC\cite{tran2019deepnc}    &   \url{https://github.com/congasix/DeepNC}  &  Python/TensorFlow & Yes                 \\ \cline{2-5} 
                                & Bacciu et al. \cite{bacciu2020edge}      &  \url{https://github.com/marcopodda/grapher}   &         Python/PyTorch & Yes          \\ \cline{2-5} 
                                 & GraphGen\cite{goyal2020graphgen}      &  \url{https://github.com/idea-iitd/graphgen}   &         Python/PyTorch & Yes          \\ \cline{2-5} 
                                                                  & GRAN\cite{liao2019efficient}          &  \url{https://github.com/lrjconan/GRAN}   &    Python/PyTorch     & Yes             \\ \cline{2-5} 
                                 & DeepGMG\cite{li2018learning}      &  \url{https://github.com/JiaxuanYou/graph-generation/blob/master/main_DeepGMG.py}   &   Python/PyTorch      & No           \\ \cline{2-5}
                                 & BiGG\cite{dai2020scalable}          &   \url{https://github.com/google-research/google-research/tree/master/bigg}  &         Python/PyTorch      & Yes    \\ \hline
\multirow{19}{*}{Autoencoder-Based}    & VGAE\cite{kipf2016variational}          &   \url{https://github.com/tkipf/gae}  &    Python/TensorFlow      & Yes          \\ \cline{2-5} 
                                 & GraphVAE\cite{simonovsky2018graphvae}     &  \url{https://github.com/JiaxuanYou/graph-generation/tree/master/baselines/graphvae}   &     Python/PyTorch     & No          \\ \cline{2-5}
                        & Graphite\cite{grover2019graphite}      &  \url{https://github.com/ermongroup/graphite}   &     Python/TensorFlow     & Yes      \\ \cline{2-5} 
                                 & NED-VAE\cite{guo2020interpretable}       &  \url{https://github.com/xguo7/NED-VAE}   &          & Yes      
                        \\ \cline{2-5} 
                        & DGVAE \cite{li2020dirichlet}        &  \url{https://github.com/xiyou3368/DGVAE}   &   Python/TensorFlow       & Yes         \\ \cline{2-5} 
                                 & JT-VAE\cite{jin2018junction}        &  \url{https://github.com/wengong-jin/icml18-jtnn}   &   Python/PyTorch       & Yes         \\ \cline{2-5} 
                                 & HierVAE\cite{jin2020hierarchical}       &  \url{https://github.com/wengong-jin/hgraph2graph}   &    Python/PyTorch      & Yes          \\ 
                                 \cline{2-5} 
                                 & MHG-VAE\cite{kajino2019molecular}       &   \url{https://github.com/ibm-research-tokyo/graph_grammar}  &    Python/PyTorch      & Yes                 
                                 \\ \cline{2-5} 
                                 & MoleculeChef\cite{bradshaw2019model}  &   \url{https://github.com/john-bradshaw/molecule-chef}  &     Python/PyTorch     & Yes          
                                 \\
                                 \cline{2-5} 
                                 & CGVAE\cite{liu2018constrained}         &   \url{https://github.com/microsoft/constrained-graph-variational-autoencoder}  &     Python/TensorFlow     & Yes          \\ \cline{2-5}
                                 & NeVAE\cite{samanta2019nevae}         &  \url{https://github.com/Networks-Learning/nevae}   &    Python/TensorFlow      & Yes         \\ \cline{2-5} 
                                                                  & Lim et al.\cite{lim2020scaffold}    &   \url{https://
github.com/jaechanglim/GGM}  &       & Yes             \\  \hline
\multirow{7}{*}{RL-Based}        & GCPN \cite{you2018graph}          &  \url{https://github.com/bowenliu16/rl_graph_generation}   &    Python/TensorFlow      & Yes          \\ \cline{2-5} 
                                 & Karimi et al. \cite{karimi2020network}    &  \url{https://github.com/Shen-Lab/Drug-Combo-Generator}   &      Python/TensorFlow    & Yes          \\ \cline{2-5} 
                                 & DeepGraphMolGen\cite{khemchandani2020deepgraphmolgen}  &  \url{https://github.com/dbkgroup/prop_gen}   &    Python/PyTorch      & Yes          \\\cline{2-5} 
                                 & MNCE-RL \cite{xu2020reinforced}  &  \url{https://github.com/Zoesgithub/MNCE-RL}   &    Python/PyTorch      & Yes          \\\cline{2-5} 
                                 & GEGL \cite{ahn2020guiding}    &  \url{https://github.com/sungsoo-ahn/genetic-expert-guided-learning}   &      Python/PyTorch    & Yes          \\  \hline
\multirow{7}{*}{Adversarial}    
                                 & NetGAN\cite{bojchevski2018netgan}        &   \url{https://github.com/danielzuegner/netgan}  &    Python/TensorFlow      & Yes         \\\cline{2-5} 
                                 & MolGAN\cite{de2018molgan}        &  \url{https://github.com/nicola-decao/MolGAN}   &    Python/TensorFlow      & Yes          \\ \cline{2-5} 
                                 & CONDGEN\cite{yang2019conditional}      & \url{https://github.com/KelestZ/CondGen}    &    Python/PyTorch      & Yes          \\ \cline{2-5} 
                                 & VJTNN + GAN \cite{jin2018learning}      & \url{https://github.com/wengong-jin/iclr19-graph2graph}    &    Python/PyTorch      & Yes          \\ \cline{2-5} 
                                 & Mol-CycleGAN\cite{maziarka2020mol}  &   \url{https://github.com/ardigen/mol-cycle-gan}  &       Python/Keras   & Yes          \\ \cline{2-5} 
                                 
                                 & Misc-GAN\cite{zhou2019misc}      &   \url{https://github.com/Leo02016/Miscgan}  &   Python/TensorFlow, Matlab       & Yes          \\  
                                 \hline
\multirow{5}{*}{Flow-Based}       & GNF\cite{liu2019graph}           &   \url{https://github.com/jliu/graph-normalizing-flows}  &     Python/TensorFlow     & Yes                 \\ \cline{2-5}
& GraphNVP \cite{madhawa2019graphnvp}           &   \url{https://github.com/Kaushalya/graph-nvp}  &     Python/Chainer-Chemistry     & Yes                 \\ \cline{2-5}
                                 & GraphAF\cite{shi2020graphaf}       &  \url{https://github.com/DeepGraphLearning/GraphAF}   &          Python/PyTorch      & Yes        \\ 
                                 \hline
\end{tabular}
\caption{\label{tab:codes} A set of publicly available source codes. O.A. = Original Authors}
\end{center}
\end{table*}

\vspace{12pt}
\begin{flushleft}
{
\justify
\bibliographystyle{unsrt}
\bibliography{IEEEabrv,references}
}
\end{flushleft}

\begin{IEEEbiography}[{\includegraphics[width=1in,height=1.25in,clip,keepaspectratio]{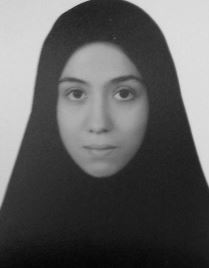}}]{Faezeh Faez} received her B.Sc. and the M.Sc. degrees in Software Engineering from Sharif University of Technology, Tehran, Iran. She is currently a Ph.D. candidate in Artificial Intelligence in the Department of Computer Engineering at Sharif University of Technology. Her current research interests include machine learning, deep learning, and deep graph generative models.
\end{IEEEbiography}

\begin{IEEEbiography}[{\includegraphics[width=1in,height=1.25in,clip,keepaspectratio]{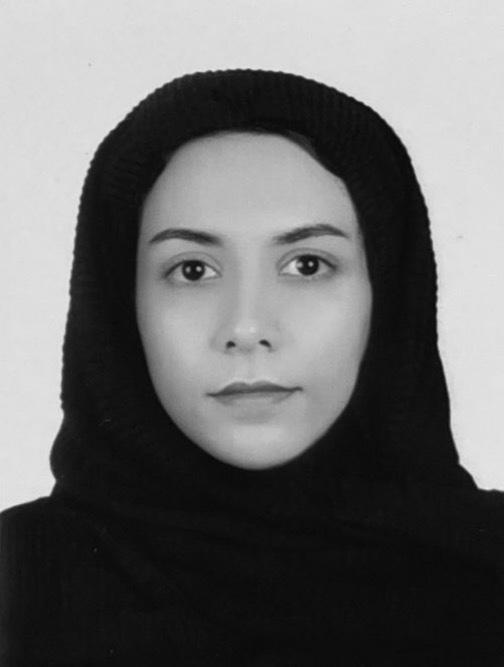}}]{Yassaman Ommi} is currently a B.Sc. student in computer science at Amirkabir University of Technology (Tehran Polytechnic), Tehran, Iran. Her current research interests include graph-based deep learning, pattern recognition, and complex networks.
\end{IEEEbiography}

\begin{IEEEbiography}[{\includegraphics[width=1in,height=1.25in,clip,keepaspectratio]{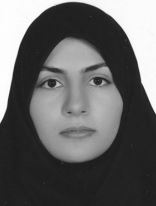}}]{Mahdieh Soleymani Baghshah} received the B.Sc., M.Sc., and Ph.D. degrees from the Department of Computer Engineering, Sharif University of Technology, Iran, in 2003, 2005, and 2010, respectively. She is an assistant professor with the Computer Engineering Department, Sharif University of Technology, Tehran, Iran. Her research interests include machine learning and deep learning.
\end{IEEEbiography}

\begin{IEEEbiography}[{\includegraphics[width=1in,height=1.25in,clip,keepaspectratio]{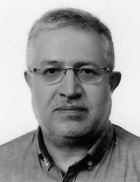}}]{Hamid R. Rabiee} (SM’07) received his BS and MS degrees (with Great Distinction) in Electrical Engineering from CSULB, Long Beach, CA (1987, 1989), his EEE degree in Electrical and Computer Engineering from USC, Los Angeles, CA (1993), and his Ph.D. in Electrical and Computer Engineering from Purdue University, West Lafayette, IN, in 1996. From 1993 to 1996 he was a Member of Technical Staff at AT\&T Bell Laboratories. From 1996 to 1999 he worked as a Senior Software Engineer at Intel Corporation. He was also with PSU, OGI and OSU universities as an adjunct professor of Electrical and Computer Engineering from 1996-2000. Since September 2000, he has joined Sharif University of Technology, Tehran, Iran. He was also a visiting professor at the Imperial College of London for the 2017-2018 academic year. He is the founder of Sharif University Advanced Information and Communication Technology Research Institute (AICT), ICT Innovation Center, Advanced Technologies Incubator (SATI), Digital Media Laboratory (DML), Mobile Value Added Services Laboratory (VASL), Bioinformatics and Computational Biology Laboratory (BCB) and Cognitive Neuroengineering Research Center. He is also a consultant and member of AI in Health Expert Group at WHO. He has been the founder of many successful High-Tech start-up companies in the field of ICT as an entrepreneur. He is currently a Professor of Computer Engineering at Sharif University of Technology, and Director of AICT, DML, and VASL. He has received numerous awards and honors for his Industrial, scientific and academic contributions, and holds three patents. His research interests include statistical machine learning, Bayesian statistics, data analytics and complex networks with applications in social networks, multimedia systems, cloud and IoT privacy, bioinformatics, and brain networks.
\end{IEEEbiography}

\EOD

\end{document}